\documentclass{article}

\usepackage{microtype}
\usepackage{graphicx}
\usepackage{subfigure}
\usepackage{booktabs} 
\usepackage{hyperref}

\usepackage[accepted]{icml2020_near_camera}

\usepackage{amsmath,amssymb,amsthm,mathrsfs,amsfonts,dsfont,tikz}

\usepackage{algorithm,algorithmic}

\usepackage{hyperref}
\hypersetup{final}

\usepackage{algorithm,algorithmic}

\usepackage{tikz}
\usepackage{etoolbox}
\newcommand{\circled}[2][]{\tikz[baseline=(char.base)]
    {\node[shape = circle, draw, inner sep = 1pt]
    (char) {\phantom{\ifblank{#1}{#2}{#1}}};
    \node at (char.center) {\makebox[0pt][c]{#2}};}}
\robustify{\circled}

\newtheorem{thm}{Theorem}
\newtheorem{prop}{Proposition}
\newtheorem{lem}{Lemma}
\newtheorem{cor}[thm]{Corollary}
\newtheorem{defi}[thm]{Definition}

\newtheorem{assumption}{Assumption}

\def\E{\mathrm{E}}

\def\Var{\text{Var}}

\def\bbR{\mathbb{R}}

\def\v{{\bf v}}
\def\z{{\bf z}}

\def\dist{\text{dist}}
\def\dom{\text{dom}}

\def\w{{\bf{w}}}

\begin{document}

\twocolumn[
\icmltitle{Stochastic Optimization for Non-convex Inf-Projection Problems}



\icmlsetsymbol{equal}{*}

\begin{icmlauthorlist}
\icmlauthor{Yan Yan}{UIOWA}
\icmlauthor{Yi Xu}{BABA}
\icmlauthor{Liijun Zhang}{NJU}
\icmlauthor{Xiaoyu Wang}{CUHK}
\icmlauthor{Tianbao Yang}{UIOWA}
\end{icmlauthorlist}

\icmlaffiliation{UIOWA}{University of Iowa}
\icmlaffiliation{BABA}{DAMO Academy, Alibaba Group}
\icmlaffiliation{NJU}{Nanjing University}
\icmlaffiliation{CUHK}{The Chinese University of Hong Kong (Shenzhen)}

\icmlcorrespondingauthor{Yan Yan}{yan-yan-2@uiowa.edu}
\icmlcorrespondingauthor{Tianbao Yang}{tianbao-yang@uiowa.edu}

\icmlkeywords{Machine Learning, ICML}

\vskip 0.3in
]



\printAffiliationsAndNotice{}  

\begin{abstract}
In this paper, we study a family of non-convex and possibly non-smooth inf-projection minimization problems, where the target objective function is equal to minimization of a joint function over another variable. This problem include difference of convex (DC) functions  and a family of bi-convex functions as special cases. We develop stochastic algorithms and establish their first-order convergence for finding a (nearly) stationary solution of the target non-convex function under different conditions of the component functions. To the best of our knowledge, this is the first work that comprehensively studies stochastic optimization of non-convex inf-projection minimization problems with provable convergence guarantee. Our algorithms enable efficient stochastic optimization of a family of non-decomposable DC functions and a family of bi-convex functions. To demonstrate the power of the proposed algorithms we consider an important application in variance-based regularization.
Experiments verify the effectiveness of our inf-projection based formulation and the proposed stochastic algorithm in comparison with previous stochastic algorithms based on the min-max formulation for achieving the same effect. 
\end{abstract}

\section{Introduction}\label{sec:introduction}
In this paper, we consider a family of non-convex and possibly non-smooth problems with the following structure
\begin{align}\label{eq:problem_setting}
\min_{x \in X} F(x) := \{g(x) + \min_{y \in \dom(h)} h(y) - \langle y, \ell(x) \rangle \} ,
\end{align}
where
$X \subseteq \mathbb{R}^{d}$ is a closed convex set,
$g: X \rightarrow \mathbb{R}$ is lower-semicontinuous,  
$h: \dom(h) \rightarrow \mathbb{R}$ is uniformly convex, 
$\ell : X \rightarrow \mathbb{R}^{m}$ is a lower-semicontinuous differentiable mapping, and
$\langle \cdot, \cdot \rangle$ is the inner product.
The requirement of uniform convexity on $h$ is to ensure the inner minimization problem is well defined and its solution is unique (cf. Section \ref{sec:pre}). 
Define $f(x, y) = g(x) + h(y) - \langle y, \ell(x)\rangle$, the objective function $F(x)$ is called the inf-projection of $f(x, y)$ in the literature.  
When $g$ is convex, depending on $\dom(h)$, the two subfamilies of above problem (\ref{eq:problem_setting}) deserve more discussion: difference of convex (DC) and bi-convex functions.

\begin{table*}[t]
	\caption{Summary of  results for finding a (nearly) $\epsilon$-stationary solution in this work under different conditions of $g$, $h$ and $\ell$. SM means smoothness, Lip. means Lipschitz continuous, Diff means differentiable, MO means monotonically increasing or decreasing for $h^*$,  CVX means convex, and UC means $p$-uniformly convex  ($p\geq 2$), $v  = 1/(p-1)$.
}
	\centering
	\label{tab:2}
	\scalebox{1}{\begin{tabular}{l|l|l|l|l|l}
			\toprule
			$g$&  $h$ ($h^*$) & $\ell$  &Alg. &Mini-Batch & Compl.\\
			\midrule
			SM& UC \& simple &SM \& Lip &MSPG (Section \ref{sec:smooth_section})&Yes&$O(1/\epsilon^{4/v})$\\
			\midrule
			SM \& CVX & UC (MO)& Diff \& Lip \& CVX &St-SPG (Section \ref{sec:practical_section})&No&$O(1/\epsilon^{4/v})$\\ 
			\midrule
		         Lip \& CVX &UC (MO) &SM \& Lip \& CVX &St-SPG (Section \ref{sec:practical_section})&No& $O(1/\epsilon^{4/v})$\\
			\bottomrule
	\end{tabular}}
\end{table*}

{\bf DC functions.} When $g$ is convex and $\dom(h) \subseteq \mathbb{R}^{m}_{+}$ and $\ell$ is convex~\footnote{$\dom(h) \subseteq \mathbb{R}^{m}_{-}$ and $\ell$ is concave can be transferred to the considered case by a variable change.}, the inf-projection minimization problem (\ref{eq:problem_setting}) is equivalent to the following DC functions,
\begin{align}\label{eq:problem_setting_dc}
\min_{x \in X} F(x) & =\Big\{  g(x) - h^*(\ell(x)) \Big\}, 
\end{align}
where $h^*$ denotes the convex conjugate function of $h$, the convexity of the second component $h^*(\ell(x))$ is following the composition rule of convexity~\citep{boyd-2004-convex}~\footnote{Note that $h^*$ is monotonically increasing  iff $\dom(h) \subseteq \mathbb{R}^{m}_{+}$.}. Minimizing DC functions has wide applications in machine learning and statistics~\cite{nitanda2017stochastic,NIPS2017_6765}. Although stochastic algorithms for DC problems have been considered recently~\cite{nitanda2017stochastic,xu2018stochastic,pmlr-v70-thi17a}, working with the inf-projection minimization~(\ref{eq:problem_setting}) is preferred 
when $h^*(\ell(x))$ is {\it non-decomposable} such that an unbiased stochastic gradient of $h^*(\ell(x))$ is not easily accessible  as that of $ \langle y, \ell(x)\rangle$ in~(\ref{eq:problem_setting}).
Inspired by this scenario, let us particularly consider an important instance variance-based regularization.
It refers to a learning paradigm that minimizes the empirical loss and its variance simultaneously, by which a better bias-variance trade-off may be achieved~\cite{maurer2009empirical}. 
To give a condensed understanding of its connection to the inf-projection formulation, we can re-formulate the problem (cf. the details and comparison with a related convex objective of \cite{namkoongnips2017variance} in Section \ref{sec:app}):
\begin{align}\label{eqn:dro}
\min_{x \in X} 
  & \frac{1}{n}\sum_{i=1}^{n} l_{i}(x) + \lambda \frac{1}{2n}\sum_{i=1}^{n} ( l_{i} (x) )^{2} - \frac{\lambda}{2} \Big( \frac{1}{n}\sum_{i=1}^{n} l_{i} (x) \Big)^{2}  ,
\end{align}
where $l_i(x): X \rightarrow \mathbb R_+$ is the loss function of a model $x$ on the $i$-th example and $\lambda>0$ is a regularization parameter. 
The above problem~(\ref{eqn:dro}) is a special case of~(\ref{eq:problem_setting_dc}) by regarding $g(x)$ as the sum of the first two terms,  
$\ell(x) = \frac{1}{n}\sum_{i=1}^nl_i(x)$ and $h^*(s) = \frac{\lambda}{2} s^2$. 
By noting the convex conjugate $- \frac{1}{2} \Big(1/n \sum_{i=1}^{n} l_{i} (x) \Big)^{2} = \min_{y\geq 0} \frac{1}{2} y^2 - \langle y, 1/n\sum_{i=1}^{n} l_{i}(x) \rangle$, 
the above problem can be viewed as a special case of~(\ref{eq:problem_setting}). 
In this way, computing a stochastic gradient of $f(x, y)$ in terms of $x$ can be done based on one sampled loss function $l_i(x)$.
It is easier than computing an unbiased stochastic gradient of $h^*(\ell(x))$ in (\ref{eqn:dro}) that requires at least two sampled loss functions.

{\bf Bi-convex functions.} When $g$ is convex and $\dom(h) \subseteq \mathbb{R}^{m}_{-}$ and $\ell$ is convex, the inf-projection minimization problem (\ref{eq:problem_setting}) reduces to minimization of a bi-convex function. 
In particular, $f(x, y)$ is convex in terms of $x$ for every fixed $y\in\dom(h)$ and $f(x, y)$ is convex in terms of $y$ for every fixed $x\in X$.  
The concerned family of bi-convex functions also find some applications in machine learning and computer vision \cite{Kumar:2010:SLL:2997189.2997322,DBLP:conf/eccv/ShahYCJSG16}.
For example, the self-paced learning method proposed by \cite{Kumar:2010:SLL:2997189.2997322} needs to solve the following bi-convex problem
$
\min_{\w \in \mathbb R^d, \v \in \{ 0, 1 \}^n } r(\w) + \sum_{i=1}^n v_i f_i(\w) - \frac{1}{K} \sum_{i=1}^n v_i  ,
$
where $r$ and $f_i$ are convex in $\w$, $v_i=0$ if $f_i(\w) > \frac{1}{K}$ and $v_i=1$ if $f_i(\w) < \frac{1}{K}$,
which can be covered by (\ref{eq:problem_setting}). 
Although deterministic optimization methods (e.g., proximal alternating linearized minimization and its variants \cite{bolte:2014:PAL:2650160.2650169,davis2016sound}) and their convergence theory have been studied for minimizing a bi-convex function~\cite{Gorski2007},
algorithms and convergence theory for stochastic optimization of a bi-convex function remains under-explored especially when we are interested in the convergence respect to the target function $F(x)$. 
A special case that belongs to both DC and Bi-convex functions is when $\ell(x) = Ax$, and $\dom(h)$ can be any convex set.

A naive idea to tackle~(\ref{eq:problem_setting}) is by alternating minimization or block coordinate descent, i.e., alternatively solving the inner minimization problem over $y$ given $x$ and then updating $x$ by certain approaches (e.g., stochastic gradient descent) \cite{bolte:2014:PAL:2650160.2650169,davis2016sound,hong2015unified,xu2013block,driggs2020spring}. 
However, this approach suffers from two issues: 
(i) solving the inner minimization might not be a trivial task (e.g., solving the inner minimization problem related to~(\ref{eqn:dro}) requires passing $n$ examples once); 
(ii) the target objective function $F(x)$ is not necessarily a smooth function or a convex function, which makes the convergence analysis challenging. 
Additionally, their convergence analysis focus on $f(x, y)$ instead of $F(x)$.
In this paper, the main question that we tackle is: how to design efficient  stochastic algorithms using simple updates for both $x$ and $y$ to enjoy a provable convergence guarantee in terms of finding a stationary point of  $F(x)$?  {\bf Our contributions} are summarized below: 
\begin{itemize}
\item First, we consider the case when $g$ and $\ell$ are  smooth but not necessarily convex and $h$ is a simple function whose proximal mapping is easy to compute. Under the condition that $\ell$ is Lipschitz continuous, we prove the convergence of mini-batch stochastic proximal gradient method (MSPG) with increasing mini-batch size that employ parallel stochastic gradient updates for $x$ and $y$, and establish the convergence rate. 

\item Second, we consider the cases when $g$ and $\ell$ are not necessarily smooth but convex, and $h$ is not necessarily a simple function (corresponding to DC and bi-convex functions). We develop an algorithmic framework that employs a suitable stochastic algorithm for solving strongly convex functions in a stagewise manner.  We analyze the convergence rates for finding a (nearly) stationary point when employing the stochastic proximal gradient (SPG) method at each stage, resulting St-SPG.  
The complexity results of our algorithms under different conditions of $g$, $h$ and $\ell$ are shown in Table~\ref{tab:2}.
\end{itemize}
The novelty and significance of our results are 
(i) this is the first work that comprehensively studies the stochastic optimization of a non-smooth non-convex inf-projection problem; 
(ii) the application of the inf-projection formulation to variance-based regularization demonstrates much faster convergence of our algorithms comparing with existing algorithms based on a min-max formulation.

\section{Preliminaries}\label{sec:pre}

Let us first present some notations.
We let $\|\cdot\|$ denote the Euclidean norm of a vector and the spectral norm of a matrix. We use $\xi$ to denote some random variable. 
Given a function $g: \bbR^{d} \rightarrow \bbR$, we denote the Fr\'echet subgradients and limiting Fr\'echet gradients by $\hat{\partial} g$ and $\partial g$ respectively, i.e., at $x$, 
$
\hat{\partial} g(x) = \{ y \in \bbR^{d} : \lim_{x \rightarrow x'} \inf \frac{ g(x) - g(x') - y^{\top} }{ \| x - x' \| } \geq 0 \}  ,
$ 
and
$
\partial g(x) = \{ y \in \bbR^{d} : \exists x_{k} \stackrel{g}{\rightarrow} x, v_{k} \in \hat{\partial g(x_{k})}, v_{k} \rightarrow v \}.
$
Here $x_{k} \stackrel{g}{\rightarrow} x$ represents $x_{k} \rightarrow x$ and $g(x_{k}) \rightarrow g(x)$.
When the function $g$ is differentiable, the subgradients ($\hat{\partial} g$ and $\partial g$ ) reduce to the standard gradient $\nabla g$. 
It is known that $ \hat{\partial} g(x) \subset \partial g(x)$, $\hat{\partial} g(x) = \{ \nabla g(x) \}$ if $g(x)$ is differential and
$\partial g(x) = \{ \nabla g(x) \}$ if $g(x)$ is continuously differential.  We denote by $\partial_{x} g(x, y) $ the partial derivative in the direction of $x$ and $\partial g(x, y) = (\partial_{x} g(x, y), \partial_{y} g(x, y))^{\top}$. In this paper, we will prove the convergence in terms of the limiting gradient. But all results can be extended to the Fr\'echet subgradients. 

Let $\nabla\ell(x)\in\mathbb R^{m\times d}$ denote the  Jacobian  matrix of the differentiable mapping $\ell(x)$. $\ell$ is said $G_\ell$-Lipschitz continuous if $\|\nabla \ell(x)\|\leq G_\ell$. 
  A differentiable function $f(\cdot)$ has $(L, v)$-H\"older continuous gradient  if $\| \nabla f(x_1) - \nabla f(x_2) \| \leq L \| x_1 - x_2 \|^{v}$ holds for some $v\in(0,1]$ and $L>0$. When $v=1$, it is known as $L$-smooth function. If $\nabla f$ is H\"older continuous,
then it holds
$f(x_1) - f(x_2) \leq \langle \nabla f(x_2), x_1 - x_2 \rangle + \frac{L}{1+v} \| x_1 - x_2 \|^{1+v}.$
A related condition is uniform convexity. A function $f(\cdot)$ is $(\varrho, p)$-uniformly convex where $p \geq 2$, if $f(x_1) - f(x_2) \geq \langle \partial f(x_2), x_1 - x_2 \rangle + \frac{\varrho}{2} \| x_1 - x_2 \|^{p}_{2} $. When $p=2$, it is known as strong convexity.  If $f$ is $(\varrho, p)$-uniformly convex, then the following inequality holds
\begin{align}\label{eq:uniformly_property}
       \varrho \| x_1 - x_2 \|^{p} 
\leq &
       \langle \partial f(x_1) - \partial f(x_2), x_1 - x_2 \rangle
       \nonumber\\
\leq &
       \| \partial f(x_1) - \partial f(x_2) \| \cdot \| x_1 - x_2 \|   .
\end{align}
It is obvious that a uniformly convex function has a unique minimizer.
If $f$ is uniformly convex, then its convex conjugate $f^*$ has H\"older continuous gradient and vice versa, which is summarized in the following lemma.

\begin{lem}\label{lemma:Holder_to_uniformly_convex}
Let $f$ be differentiable and 
$\nabla f$ be $(L, v)$-H\"older continuous where $v \in (0, 1]$.
Then $f^*$ is $(\varrho, p)$-uniformly convex with 
$p = 1 + \frac{1}{v}$ 
and 
$\varrho = \frac{ 2 v }{ 1 + v } (1/L )^{\frac{1}{v}} $. If $f$ is $(\varrho, p)$-uniformly convex, then $f^*$ has $(L, v)$-H\"older continuous gradient with $L = (1/\varrho)^{1/(p-1)}$ and $v = 1/(p-1)$. 
\end{lem}

Next we discuss the convergence measure for the considered inf-projection problem.
Let $f_h (x)= \min_{y}h(y) - y^{\top}\ell(x)$. 
If $h$ is uniformly convex, let $y^*(x) = \arg\min_y h(y) - y^\top \ell(x)$ denote the unique minimizer. 
In this way, under a regularity condition that $h(y) - y^{\top}\ell(x)$ is level-bounded in $y$ uniformly in $x$, then Theorem 10.58 of \cite{rockafellar2009variational} implies $\partial f_h(x) = -\nabla \ell(x)^{\top} y^*(x)$. 
Then $\partial F(x) = \partial g(x) - \nabla \ell(x)^{\top} y^*(x)$ under the smoothness or convexity condition of $g$, which allows us to connect $\partial F(x)$ by $\partial g(x) - \nabla \ell(x)^{\top} y^*(x)$.
In this paper, we aim to find a solution that is $\epsilon$-stationary or nearly $\epsilon$-stationary of $F$, which are defined as follows.
\begin{defi}\label{def:epsilon_critical}
A solution $x$ satisfying $\dist(0, \partial F(x))\leq \epsilon$  is called an $\epsilon$-stationary point of $F$.
A solution  $x$ is called a nearly $\epsilon$-stationary if there exists $z$ and a constant $c>0$ such that $\| z - x \| \leq c\epsilon$ and  $\dist(0, \partial F(z)) \leq \epsilon$. 
\end{defi}
Particularly, nearly stationarity has been used to measure the convergence for non-smooth non-convex optimization in the literature~\cite{davis2017proximally,sgdweakly18,modelweakly18,chen18stagewise,xu2018stochastic}.

Before ending this section, we state basic assumptions below.  For simplicity, here all variance bounds are denoted by $\sigma^2$. Additional conditions regarding $g$, $h$ and $\ell$ are presented in individual theorems.  
\begin{assumption}\label{assumption:loss_function}
For the problem~(\ref{eq:problem_setting}) we assume: \\
 (i)  $h^*$ has  $(L_{h_*}, v)$-H\"older continuous gradient, and $\ell$ is continuously differentiable; \\
 (ii)  Let $\partial g(x; \xi_g)$ denote a stochastic gradient of $g(x)$. If $g(x)$ is smooth, assume $\E[\|\partial g(x; \xi_g) - \partial g(x)\|^2]\leq \sigma^2$, otherwise assume  $\E[\|\partial g(x; \xi)\|^2]\leq \sigma^2$ for $x\in X$; \\
 (iii) Let $\ell(x; \xi_\ell)$ denote a stochastic version  of $\ell(x)$ and assume $\E[\|\ell(x; \xi_\ell) - \ell(x)\|^2]\leq\sigma^2$. If $\ell(x)$ is smooth, assume $\E[\|\nabla \ell(x; \xi_\ell) - \nabla \ell(x)\|^2]\leq \sigma^2$, otherwise assume  $\E[\|\nabla \ell(x; \xi_\ell)\|^2]\leq \sigma^2$ for $x\in X$; \\
 (iv)  Let $\partial h(y; \xi_h)$ denote a stochastic gradient of $h(y)$ and assume  $\E[\|\partial h(y;\xi_h)\|^2]\leq \sigma^2$ for $y\in\dom(h)$; \\
 (v) $\max_{x\in X,y\in \dom(h)} f(x,y) - \min_{x\in X, y\in\dom(h)} f(x,y) \leq M$.
\end{assumption}

\section{Mini-batch Stochastic Gradient Methods For Smooth Functions}
\label{sec:smooth_section}

\begin{algorithm}[t]
    \caption{MSPG}\label{alg:simple_minmin_sgd_one_loop}
\begin{algorithmic}[1]
    \STATE \textbf{Input}: initialized $x_{1}, y_{1}$.

    \FOR{$t=1,\ldots, T$}
      \STATE Compute mini-batch stochastic partial gradients $\widetilde\nabla_x f^{(t)}_0=\frac{1}{m_t}\sum_{i=1}^{m_t} \nabla_{x} f_0(x_{t}, y_{t}; \xi_{i} )$ and $\widetilde\nabla_y f^{(t)}_0=\frac{1}{m_t}\sum_{i=1}^{m_t}\nabla_{y} f_0(x_{t}, y_{t}; \xi_{i} )$

      \STATE $ x_{t+1} =\Pi_{X}[x_{t} - \eta \widetilde\nabla_{x} f^{(t)}_0]$
        
      \STATE $y_{t+1} =P_{\eta h} [y_{t} - \eta\widetilde\nabla_{y} f^{(t)}_0]$

    \ENDFOR
    
    \STATE \textbf{Output}: $w_\tau=(x_{\tau},y_{\tau})$, where $\tau\in\{1,\ldots, T\}$ is randomly sampled 
\end{algorithmic}
\end{algorithm}

In this section, we consider the case when $g$ and $\ell$ are smooth functions but not necessarily convex. 
Please note that the target function $F$ is still not necessarily smooth and is non-convex.
We assume $h$ is simple such that its proximal mapping defined by $P_{\eta h}[\hat y] = \arg\min_{y}h(y) + \frac{1}{2\eta}\|y -\hat y\|^2$ is easy to compute. 
Let $f_0(x, y) = g(x) - y^{\top}\ell(x)$. 
The key idea of the our first algorithm is that we treat $f(x, y) = f_0(x, y) + h(y) + I_{X}(x)$ as a function of the joint variable $w = (x, y)$, which consists of a smooth component $f_0$ and a non-smooth component $h(y) + I_{X}(x)$. 
Hence, we can employ mini-batch stochastic proximal gradient (MSPG) method to minimize $f(x, y)$ based on stochastic gradients of $f_0(w)$ denoted by $\nabla f_0(w; \xi)$ for a random variable $\xi$.  
The detailed steps of MSPG are shown in Algorithm~\ref{alg:simple_minmin_sgd_one_loop}. 
At each iteration, stochastic partial gradients w.r.t. $x$ and $y$ are computed and used for updating.


Although the convergence of MSPG for $f(w)$ has been considered in literature of composite optimization~\cite{DBLP:journals/mp/GhadimiLZ16} or alternating minimization \cite{hong2015unified,xu2013block,driggs2020spring},
there is still a gap when applying the existing convergence result, since we are interested in the convergence analysis of $\dist(0, \partial F(x))$, rather than $f_0(w)$.
In the following, we fill this gap by four main steps.
In brief, first, we establish the joint smoothness of $f_0$ in $(x, y)$ by Lemma \ref{lemma:joint_L_smooth1}.
Then, based on Lemma \ref{lemma:joint_L_smooth1}, we derive the convergence of $\dist(0, \partial f(w))$ in Proposition \ref{prop:1}.
Next, Lemma \ref{lem:for:prop2} connects $\dist(0, \partial f(w))$ to $\dist(0, \partial F(x))$.
Finally, the convergence of $\dist(0, \partial F(x))$ is achieved (Theorem \ref{thm:smooth_one_loop}).

\begin{lem}\label{lemma:joint_L_smooth1}
Suppose $g(x)$ is $L_{g}$ smooth, $\ell$ is $G_{\ell}$-Lipschitz continuous and $L_{\ell}$-smooth, and $\max_{y\in \dom(h)} \| y \| \leq D_{y}$.
Then $f_0(x, y)$ is smooth over $(x, y)\in X\times \dom(h)$, i.e., 
\begin{align*}
& 
\| \nabla_{x} f_0(x, y) - \nabla_{x} f_0(x', y') \|_{2}^{2}
\\
& ~~~~~~~~
+ \| \nabla_{y} f_0(x, y) - \nabla_{y} f_0(x', y') \|_{2}^{2}  
\\
& \leq
L^{2} ( \| x - x' \|_{2}^{2} + \| y - y' \|_{2}^{2} )   ,
\end{align*}
where $L = \sqrt{ \max(2L_{g}^{2} + 4 L_{\ell}^{2} D_{y}^{2} + G_{\ell}^{2} ,  4G_{\ell}^{2} ) }$.
\end{lem}

Based on the joint smoothness of $f_0$ in $(x, y)$, we can establish the convergence of MSPG in terms of $\dist(0, \partial f(x_\tau, y_\tau))$ in the following proposition. 
Note that this convergence result in terms  of $\dist(0, \partial f(x_\tau, y_\tau))$ is stronger than that in~\cite{DBLP:journals/mp/GhadimiLZ16} in terms of proximal gradient, which follows the analysis in~\cite{DBLP:journals/corr/abs/1902.07672}.  

\begin{prop}\label{prop:1}Under the same conditions as in Lemma~\ref{lemma:joint_L_smooth1} and suppose the stochastic gradient has bounded variance $\E[\|\nabla f_0(w; \xi) - \nabla f_0(w)\|^2]\leq \sigma_0^2$, run MSPG with $\eta = \frac{c}{L}$ ($0<c<\frac{1}{2}$) and a sequence of mini-batch sizes $m_t = b(t+1)$ for $t = 0,\dots, T-1$, where $b>0$ is a constant, then the output $w_\tau$ of Algorithm~\ref{alg:simple_minmin_sgd_one_loop} satisfies
\begin{align*}
\E[\text{dist}(0, \partial f(w_{\tau}))^2]\leq \frac{c_1\sigma_0^2 (\log(T)+1)}{bT} +\frac{c_2\Delta}{\eta T},
\end{align*}
where  $c_1 =  \frac{2c(1 -2c)+2}{c(1-2c)}$ and $c_2 = \frac{6-4c}{1-2c}$.
\end{prop}
The next lemma establishes the relation between $\dist(0, \partial f(w_\tau))$ and $\dist(0, \partial F(x_\tau))$, 
allowing us to bridge the convergence of $\dist(0, \partial F(x_\tau))$ by employing that of $\dist(0, \partial f(w_\tau))$.

\begin{lem}\label{lemma:asymptotic_one_loop}
Under the same conditions as in Lemma~\ref{lemma:joint_L_smooth1} and $h^*$ has $(L_{h^*}, v)$-H\"older continuous gradient.
Then for any $(\tilde x, \tilde y)\in X\times\dom(h)$, we have
\begin{align*}
&
\dist(0, \partial F(\tilde{x}))
\leq 
\| \nabla_{x} f(\tilde{x}, \tilde{y})  \|_{2}
\\
& ~~~~~~~~~~~
+ G_{\ell} \Big( \frac{ (1+v) }{ 2v } \Big)^{v} L_{h^*} \dist(0, \partial_{y} f(\tilde{x}, \tilde{y}))^{v}  .
\end{align*}
\end{lem}

Finally, combining the above results, we can state the main result in this section regarding the convergence of MSPG in terms of the concerned $\dist(0, \partial F(x_\tau))$ as follows.
\begin{thm}\label{thm:smooth_one_loop}
Suppose  the same conditions as in Lemma~\ref{lemma:joint_L_smooth1}  and Assumption~\ref{assumption:loss_function} hold. 
Algorithm \ref{alg:simple_minmin_sgd_one_loop} guarantees that $\E[\dist(0, \partial F(x_\tau))^2]\leq O(1/T^v)$. 
To ensure $\E[\dist(0, \partial F(x_\tau))] \leq \epsilon$,
we can set $T=O(1/\epsilon^{2/v})$.  
The total complexity is $O(1/\epsilon^{ 4 / v }  )$.
\end{thm}

\section{Stochastic Algorithms for Non-Smooth Functions}
\label{sec:practical_section}

In this section, we consider the case when $g$ or $\ell$ are not necessarily smooth but are convex. We also assume $h^*$ is monotonic, i.e., $\dom(h)\subseteq\mathbb R^m_+$ or $\dom(h)\subseteq\mathbb R^m_-$. In the former case, the objective function  belongs to {\bf DC functions}, and in the latter case the objective function belongs to {\bf Bi-Convex} functions.  Please note that the target function $F$ is still not necessarily convex and is non-smooth. The proposed algorithm is inspired by the stagewise stochastic DC algorithm proposed in~\cite{xu2018stochastic} but with some major changes. 
Let us first briefly discuss the main idea and logic behind the proposed algorithm. 
There are two difficulties that we need to tackle: 
(i) non-smoothness and non-convexity in terms of $x$, 
(ii) minimization over $y$.

To tackle the first issue, let us assume the optimal solution $y^*(x) = \arg\min_y h(y) - y^\top \ell(x)$ given $x$ is available. Then the problem regarding $x$ becomes: 
\begin{align}\label{eqn:xpro}
\min_{x\in X} g(x) - {y^*(x)}^{\top}\ell(x)
\end{align}
When $\dom(h) \subseteq\mathbb R^m_+$ (corresponding to a DC function), the above problem is still non-convex. In order to obtain a provable convergence guarantee, we consider the following  strongly convex problem from some $\gamma>0$ and $x_0\in X$, whose objective function is an upper bound of the function in~(\ref{eqn:xpro}) at $x_0$: 
\begin{align}\label{eqn:fix}
P(x_0) = &
\arg\min_{x\in X} \Big\{ g(x) - {y^*_{x_0}}^{\top}(\ell(x_0)
\nonumber\\
&
~~~~~~~
+ \nabla \ell(x_0)(x - x_0)) + \frac{\gamma}{2}\|x - x_0\|^2 \Big\} .
\end{align}
Note $P(x_0)$ is  uniquely defined due to strong convexity. If $x_0 = P(x_0)$ it can be shown that $x_0$ is the critical point of $F(x)$, i.e., $0\in \partial F(x_0) = \partial g(x_0) - \nabla \ell(x_0)^{\top}y^*(x_0)$.  
Then we can iteratively solve the fixed-point problem $x = P(x)$ until it converges.

When $\dom(h) \subseteq\mathbb R^m_-$ (corresponding to a Bi-convex function), we can simply consider the following strongly convex problem:
\begin{align*}
P(x_0) = \arg\min_{x\in X} g(x) - {y^*(x_0)}^{\top} \ell(x) + \frac{\gamma}{2}\|x - x_0\|^2.
\end{align*}
A remaining issue in the above approach is that $y^*(x_0)$ is assumed available, which is related to the second issue mentioned above. 
It may not be easy to obtain an exact minimizer $y^*(x_0)$ given a $x_0$.
To this end, we can employ an iterative stochastic algorithm to optimize $\min_{y} h(y) - y^{\top}\ell(x_0)$ approximately given $x_0$, and obtain an inexact solution $\hat y(x_0)$ such that $h(\hat y(x_0)) - \hat y(x_0)^{\top}\ell(x_0) - h(y^*(x_0)) - y^*(x_0)^{\top}\ell(x_0) \leq \varepsilon$ for some approximation error $\varepsilon$.
Then, we combine these two pieces together, i.e., replacing $y^*(x_0)$ in the definition of $P(x_0)$ with $\hat y(x_0)$, and employing a stochastic algorithm to solve the fixed-point equation by $x \leftarrow\hat P(x)$, where $\hat P(x)$ is an approximation of $P(x)$.
Therefore, we have two sources of approximation error --- one from using $\hat y_x$ instead of $y^*(x)$ and another one from solving the minimization problem of $x$ inexactly. 
Our analysis is to show that with well-controlled approximation error, we can still achieve provable convergence guarantee.

For the sake of presentation, let us first introduce some important notations by considering different conditions of DC and bi-convex functions.
For the $k$-th stage of St-SPG, define 
\begin{align*}
f^k_x(x) = & g(x) - y_k^{\top}(\ell(x_k) + \nabla \ell(x_k)(x - x_k))  ,
\\
& \qquad\qquad\qquad\qquad \text{~for~} \dom(h) \subseteq \mathbb R^m_+   ,
\end{align*}
and 
\begin{align*}
f^k_x(x) = & g(x) - y_k^{\top}\ell(x) 
,
\text{~for~} \dom(h)\subseteq\mathbb R^m_-  .
\end{align*}
A stochastic gradient of $f^k_x(x)$ can be computed by $\partial g(x; \xi_g) - \nabla \ell(x_k; \xi_\ell)^{\top}y_k$ for $\dom(h) \subseteq \mathbb R^m_+$ or $\partial g(x; \xi_g) - \nabla \ell(x; \xi_\ell)^{\top}y_k$ 
for $\dom(h)\subseteq\mathbb R^m_-$. 
For both conditions, let
\begin{align*}
    &
    f^k_y(y) = h(y) - y^{\top}\ell(x_{k+1})  ,
    \\ &
    R^k_x(x) = \frac{\gamma}{2}\|x - x_k\|^2  ,~~
    R^k_y(y) = \frac{\mu}{2}\|y - y_k\|^2   .
\end{align*}
A stochastic gradient of $f^k_y(y)$ can be computed by $\partial h(y; \xi_h) - \ell(x_{k+1}; \xi_\ell')$, where $\xi_g, \xi_\ell, \xi_h, \xi_\ell'$ denote independent random variables.

The proposed algorithm is shown in Algorithm~\ref{alg:sgd_two_loop} named St-SPG, which employs SPG in Algorithm~\ref{alg:SPG} to solve the subproblems of $x$ and $y$ in a stagewise manner. 
$x$ and $y$ share the same update method SPG, so we can summarize it in general notations.
To this end, let us consider the convergence of SPG for solving $H(z) = f(z) + R(z)$, where $f(z)$ is a convex function and $R(z) = \frac{\gamma}{2}\|z-z_1\|^2$ is a strongly convex function.  Its convergence has been considered in many previous works. 
Here, we adopt the results derived in~\cite{xu2018stochastic} to establish the convergence of St-SPG under different conditions of $g$ and $\ell$ as follows.
 
\begin{prop}\label{thm:inner_loop_x}
Let  $H(z) = f(z)+R(z)$ where $R(z)=\frac{\gamma}{2}\|z-z_1\|^2$ is $\gamma$-strongly convex. If $f(z)$ is $L$-smooth and $\E[|\nabla f(z; \xi)  -\nabla f(z)|^2]\leq \sigma^2$ and  $\gamma\geq 3L$, then by setting $\eta_t = 3/(\gamma(t+1))$ SPG  guarantees that 
\begin{align*}
\E[H(\hat z_T) - H(z_*)]\leq \frac{4\gamma\|z_* - z_1\|^2}{3T(T+3)} + \frac{6\sigma^2}{(T+3)\gamma}.
\end{align*}
If $f$ is non-smooth with  $\E[|\nabla f(z; \xi)\|^2]\leq \sigma^2$, then by setting $\eta_t = 4/(\gamma t)$  SPG guarantees that 
\begin{align*}
\E\bigg[H(\hat z_T) - H(z_*)\bigg]\leq  \frac{\gamma\|z_* - z_1\|^2}{4T(T+1)} +  \frac{17\sigma^2}{\gamma (T+1)},
\end{align*}
where $z_* = \arg\min_{z\in\Omega}H(z)$. 
\end{prop}

\begin{algorithm}[t]
    \caption{St-SPG }\label{alg:sgd_two_loop}
\begin{algorithmic}[1]
    \STATE Initialize $x_{1}\in X$, $y_{1}\in \dom(h)$ 
\STATE Set a sequence of integers $T^x_k$, $T^y_k$ and numbers $\gamma, \mu$

    \FOR{$k=1, \ldots, K$}

       \STATE $x_{k+1} = \text{SPG}(f^k_x,  R^k_x, x_{k}, X, T^x_k, \gamma)$

       \STATE $y_{k+1} = \text{SPG}(f^k_y,  R^k_y, y_k, \dom(h), T^y_k, \mu)$
    \ENDFOR
\end{algorithmic}
\end{algorithm}
\begin{algorithm}[t]
    \caption{SPG($f, R, z_1, \Omega, T, \gamma$)}\label{alg:SPG}
\begin{algorithmic}[1]
    \STATE Set $\eta_t$ according to $\gamma$
    \FOR{$t=1, \ldots, T$}

      \STATE $z_{t+1} = \arg\min_{z \in \Omega} \partial f(z_t; \xi_t)^{\top}z  + R(z)   
                                 + \frac{1 }{2\eta_t} \|z - z_t\|^2$

    \ENDFOR
    
    \STATE \textbf{Output}  $\hat{z}_{T} =  \frac{\sum_{t=1}^{T} t z_{t} }{\sum_{t=1}^{T} t} $
    
\end{algorithmic}
\end{algorithm}

With the above proposition, we can apply the above convergence guarantee of SPG for $f^k_x(x) + R_x^k(x)$ and $f^k_y(y) + R_y^k(y)$.
Then define $v_k$ and $u_k$ as the optimal solutions to the subproblems of $x$ and $y$ at the $k$-th stage, respectively:
\begin{align*}
v_k = & \arg\min_{x\in X}f^k_x(x) + R^k_x(x), 
\\
u_k = & \arg\min_{y} f^k_y(y) + R^k_y(y)   .
\end{align*}
We can establish the following result regarding the convergence of St-SPG related to fixed-point convergence ($x_{\tau+1} - x_\tau$), and also the minimization error of $P(x)$, i.e., $\|x_{\tau+1} - v_\tau\|$, for a randomly sampled index $\tau\in\{1,\ldots, K\}$.
We have boundedness assumptions on $y$ and $\ell$ below to guarantee the boundedness of the second moment of stochastic gradients, which can be implied by assuming the domain $X$ is a compact set and $\dom(h)$ is bounded.

\begin{thm}\label{thm:outer_loop} 
Suppose Assumption~\ref{assumption:loss_function} holds, and $\max(\|y_k\|^2, \E[\|\ell(x_{k+1}; \xi)\|^2])\leq D^2$ for $k\in\{1, \ldots, K\}$. 
There exists a constant $G=17 \max\{2\sigma^2 +2 D^2\sigma^2, 2\sigma^2 +2 D^2 \}$, and for any constants $\gamma>0, \mu>0, \alpha\geq 1$  Algorithm \ref{alg:sgd_two_loop}  with $T_k^y  = k/\gamma + 1$, $T^t_k = k/\mu + 1$ guarantees that the following inequalities hold:
\begin{align*}
\frac{1}{2}\E[  \| x_{\tau+1} -v_\tau \|^{2}]
\leq &
\E[  \| x_{\tau} -v_\tau \|^{2}  + \| x_{\tau+1} -x_\tau \|^{2}]
\\
\leq &
     \frac{ 4(M+2G^2) (\alpha + 1) }{ \gamma K }  
\\
\frac{1}{2}\E[  \| y_{\tau+1} -u_\tau \|^{2}]
\leq &
\E[  \| y_{\tau} -u_\tau \|^{2}  + \| y_{\tau+1} -y_\tau \|^{2}]
\nonumber\\
\leq &
       \frac{ 4(M+2G^2) (\alpha + 1) }{ \mu K } 
\end{align*}
for $\tau$ sampled by $P(\tau = k) = \frac{ k^\alpha }{\sum_{s=1}^{K} {s^\alpha }}$. 
\end{thm}

The lemma below connects $\| \nabla F(x_k) \|$ (or $\dist(0, F(x_k))$) to the quantities in Theorem \ref{thm:outer_loop}, by which we can derive the convergence of (nearly) stationary point.

\begin{lem}\label{lem:epsilon_critical_smooth_g} 
Suppose $g$ is $L_g$-smooth, and $\ell$ is $G_\ell$-Lipschitz continuous. 
Then for any $k$ we have
\begin{align*}
\| \nabla F(x_{k}) \|
\leq &
(\gamma + L_g) \|  x_{k}  - v_{k}\|   + G_{\ell} \| y_{k} - u_{k} \|
\\
&
+ G_{\ell} \mu^{v} \Big( \frac{1+v}{2v} \Big)^{v} L_{h^*} \| u_{k} - y_{k} \|^{v} 
\\
&
+ G_{\ell}^{v+1} \Big( \frac{1+v}{2v} \Big)^{v} L_{h^*} \| x_{k+1} - x_{k} \|^{v}
          .
\end{align*}
Suppose $g$ is non-smooth, and $\ell$ is $G_\ell$-Lipschitz continuous and $L_\ell$-smooth and $\max_{y\in\dom(h)}\|y\|\leq D$, then for  any $k$ we have
\begin{align*}
&
\dist(0, \partial F(v_{k}))
\leq  
(\gamma + DL_\ell) \|x_{k} -  v_{k}  \|
+ G_{\ell} \|  y_{k}  - u_{k}\|
\\
& ~~
+ G_{\ell} \Big( \frac{1+v}{2v} \Big)^{v} L_{h^*} \Big( \mu \| y_{k}  - u_{k} \| + G_{\ell} \| x_{k+1} - z_{k} \|  \Big)^{v}.
\end{align*}
\end{lem}

Combining Lemma~\ref{lem:epsilon_critical_smooth_g} and Theorem~\ref{thm:outer_loop}, we have the following corollaries regarding the convergence of St-SPG under different conditions of $g$ and $\ell$. 
\begin{cor}\label{cor:complexity_smooth_g_unknown_v}
Suppose $g$ is $L_g$-smooth and $\ell$ is $G_\ell$-Lipschitz continuous and both are convex. Under the same conditions as in Theorem~\ref{thm:outer_loop}, we have 
$\E[\dist(0, \nabla F(x_\tau))] \leq \epsilon$ after $K= O(\epsilon^{-\frac{ 2 }{ v } } )$ stages.
Therefore, the total iteration complexity is 
$
\sum_{k=1}^{K}(T_k^x+ T_{k}^{y}) = O( \epsilon^{-\frac{ 4 }{ v }  } )  .
$
\end{cor}

\begin{cor}\label{cor:complexity_nonsmooth_g_unknown_v}
Suppose $g$ is non-smooth and convex, $\ell$ is $G_\ell$-Lipschitz continuous and $L_\ell$-smooth and convex, and $\max_{y\in\dom(h)}\|y\|\leq D$. Under the same conditions as in Theorem~\ref{thm:outer_loop},  we have 
$\E[\dist(0, \nabla F(v_\tau))] \leq \epsilon$ and $\E[\|x_\tau - v_\tau\|]\leq O(\epsilon^{1/v})$ after $K= O(\epsilon^{-\frac{ 2 }{ v } } )$ stages.
Therefore, the total iteration complexity is 
$
\sum_{k=1}^{K}(T_k^x+ T_{k}^{y}) = O( \epsilon^{-\frac{ 4 }{ v } }  )  .
$
\end{cor}

{\bf Remark: } 
Our algorithms enjoy the same iteration complexity of that  in \cite{xu2018stochastic} for DC functions when $v$ is unknown or $v=1$,
but we do not assume a stochastic gradient of  $h^*(\ell(x))$ is easily computed. It is also notable that St-SPG doest not need the knowledge of $v$ to run.

Finally, we would like to mention that the SPG algorithm for solving subproblems in Algorithm~\ref{alg:sgd_two_loop} can be replaced by other suitable stochastic optimization algorithms for solving a strongly convex problem similar to the developments in~\cite{xu2018stochastic} for minimizing DC functions. For example, one can use adaptive stochastic gradient methods in order to enjoy an adaptive convergence, and one can use variance reduction methods if the involved functions are smooth and have a finite-sum structure to achieve an improved convergence.

\section{Application for Variance Regularization}\label{sec:app}

\begin{table}[h]
\centering
\caption{Data statistics.}
\label{tab:data_stats}
\begin{small}
\begin{tabular}{c|c|c|c}
\hline
Datasets     &    \#Examples    &    \#Features    &   \#pos:\#neg     \\\hline
a9a          &     32,561       &        123       &     0.3172:1      \\
covtype      &     581,012      &        54        &     1.0509:1      \\
RCV1         &     697,641      &       47,236     &     1.1033:1      \\
URL          &    2,396,130     &    3,231,961     &     0.4939:1      \\
\hline
\end{tabular}
\end{small}
\end{table}

In this section, we consider the application of the proposed algorithms for variance-based regularization in machine learning. 
Let $l(\theta, \z)\in\mathbb R^+$ denote a loss of model $\theta\in\Theta$ on a random data $\z$. A fundamental task in machine learning is to minimize the expected risk $R(\theta) = \E_\z[l(\theta, \z)]$. However, in practice one has to find an approximate model based on sampled data $\mathcal S_n = \{\z_1,\ldots, \z_n\}$. An advanced learning theory according to  Bennett's inequality bounds the expected risk by~\cite{maurer2009empirical}:
\begin{align}\label{eq:erm_confidence_bound}
  R(\theta) \leq  \frac{1}{n}\sum_{i=1}^nl(\theta, \z_n) + c_1 \sqrt{ \frac{ \Var(\ell(\theta, \z )) }{ n } } + \frac{c_2}{n}  ,
\end{align}
where $c_1$ and $c_2$ are  constants. This motivates the variance-based regularization approach~\cite{maurer2009empirical}: 
\begin{align}\label{eq:erm_variance_regularized}
  \min_{\theta \in \Theta} \frac{1}{n}\sum_{i=1}^nl(\theta, \z_n)+ \lambda \sqrt{ \frac{ \Var_{n} (\theta, \mathcal S_n) }{ n } }  ,
\end{align}
where $\Var_{n} (\theta,\mathcal S_n) = \frac{1}{n} \sum_{i=1}^{n} ( \ell(\theta, \z_i) - \bar l_n(\theta) ] )^{2}$ is the empirical variance of loss, $\bar l_n(\theta)$ is the average of empirical loss, and $\lambda>0$ is a regularization parameter.

However, the above formulation does not favor efficient stochastic algorithms. To tackle the optimization problem for variance-based regularization, \cite{namkoongnips2017variance} proposed a min-max formulation based on {\it distributionally robust optimization}, given below and proposed  stochastic algorithms for solving the resulting min-max formulation when the loss function is convex~\citep{DBLP:conf/nips/NamkoongD16},
\begin{align}
\label{eq:variance_regularized_robust}
\min_{\theta \in \Theta} \max_{P\in\Delta_n} \{\sum_{i=1}^n P_i\ell(\theta, X_i) ] : D_{\phi}(P || \hat{P}_{n}) \leq \rho \}  ,
\end{align}
where $\rho > 0$ is a hyper-parameter, $\Delta_n =\{P\in\mathbb R^n; P\geq 0, \sum_{i=1}^nP_i = 1\}$, $\hat P_n = (1/n, \ldots, 1/n)$, and $D_{\phi}(P||Q) = \int \phi(\frac{dP}{dQ}) dQ$ is called the $\phi$-divergence based on $\phi(t) = \frac{1}{2}(t-1)^2$. 
The  min-max formulation is convex and concave when the loss function is convex. Nevertheless, the stochastic optimization algorithms proposed for solving the min-max formulation are not scalable. The reason is that  it introduces an $n$-dimensional dual variable $P$ that is restricted on a probability simplex. As a result, the per-iteration cost could be dominated by updating the dual variable that scales as $O(n)$, which is prohibitive when the training set is large. Although one can use a special structure and a stochastic coordinate update on $P$ to reduce the per-iteration cost to $O(\log(n) )$~\citep{DBLP:conf/nips/NamkoongD16}, the iteration complexity could be still blowed up by a factor up to $n$ due to the variance in the stochastic gradient on $P$.

As a potential solution to addressing the scalability issue, we consider the following reformulation: 
\begin{align}
\label{eq:svp_youngs}
     F(\theta)
= &   
     \frac{1}{n}\sum_{i=1}^nl(\theta, \z_n)+ \lambda \sqrt{ \frac{ \Var_{n}(\theta, \mathcal S_n) }{ n } }    
     \nonumber\\
= &
     \min_{\alpha > 0} \frac{1}{n}\sum_{i=1}^nl(\theta, \z_n) + \lambda \bigg( \frac{ \Var_{n}(\theta, \mathcal S_n) }{ 2 \alpha } + \frac{ \alpha }{ 2 n } \bigg) 
     \nonumber\\
= & 
     \min_{\alpha > 0} \frac{1}{n}\sum_{i=1}^nl(\theta, \z_n)
     + \lambda \Big( \frac{\alpha}{2n}  
     \nonumber\\
  & ~~~   
     + \frac{ \frac{1}{n} \sum_{i=1}^{n} ( \ell(\theta, \z_{i}))^2 - (\E_i[ \ell(\theta, z_i) ] )^{2} }{ 2 \alpha } 
     \Big).
\end{align}
In practice, one usually needs to tune the regularization parameter $\lambda$ in order to achieve the best performance. As a result, we can further simplify the problem  by absorbing $\alpha$ into the regularization parameter $\lambda$ and end up with the following formulation by noting $-\frac{1}{2}s^2 = \max_{y\geq 0}  \frac{1}{2}y^2 - ys$  for $s\geq 0$:
\begin{align}\label{eq:dro_inf_projection_formulation}
\min_{\theta \in \Theta} 
&
\frac{1}{n} \sum_{i=1}^{n} l(\theta, \z_{i}) 
+ \lambda \frac{1}{2n}\sum_{i=1}^{n} ( l(\theta, \z_{i}))^2 
\nonumber\\
& 
+ \lambda \{ \min_{y\geq 0}\frac{1}{2}y^2 -  y\frac{1}{n}\sum_{i=1}^{n} l(\theta, \z_{i})  \} .
\end{align}
It is notable that the above formulation only introduces one additional scalable variable $y\in\mathbb R^+$, though the problem might become a non-convex problem of $\theta$. However, when the loss function $l(\theta, \z)$ itself is a non-convex function, the min-max formulation~(\ref{eq:variance_regularized_robust}) also losses its convexity, which makes our inf-projection formulation more favorable.

\begin{figure*}[t]
\centering
{\includegraphics[scale=.226]{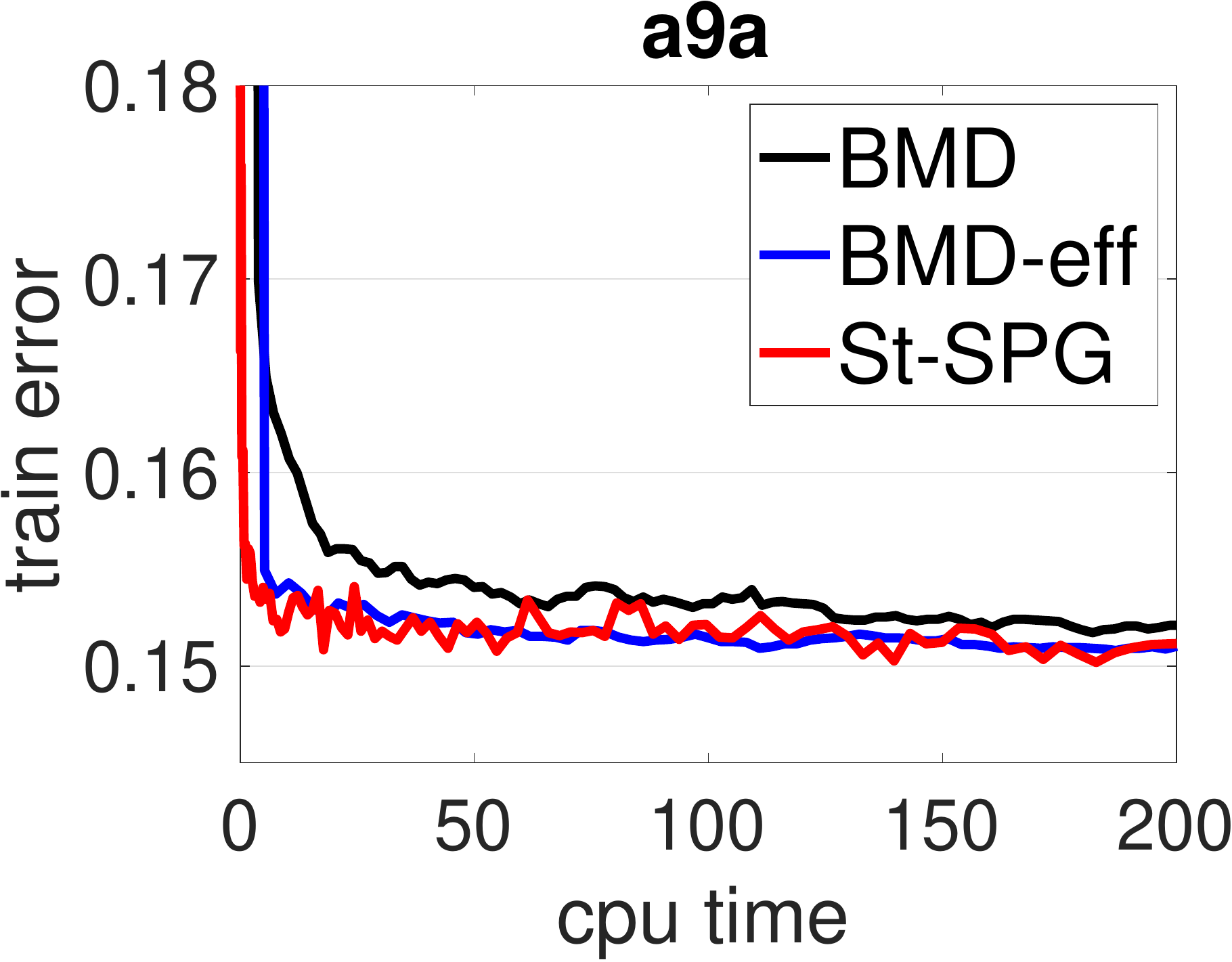}}
{\includegraphics[scale=.226]{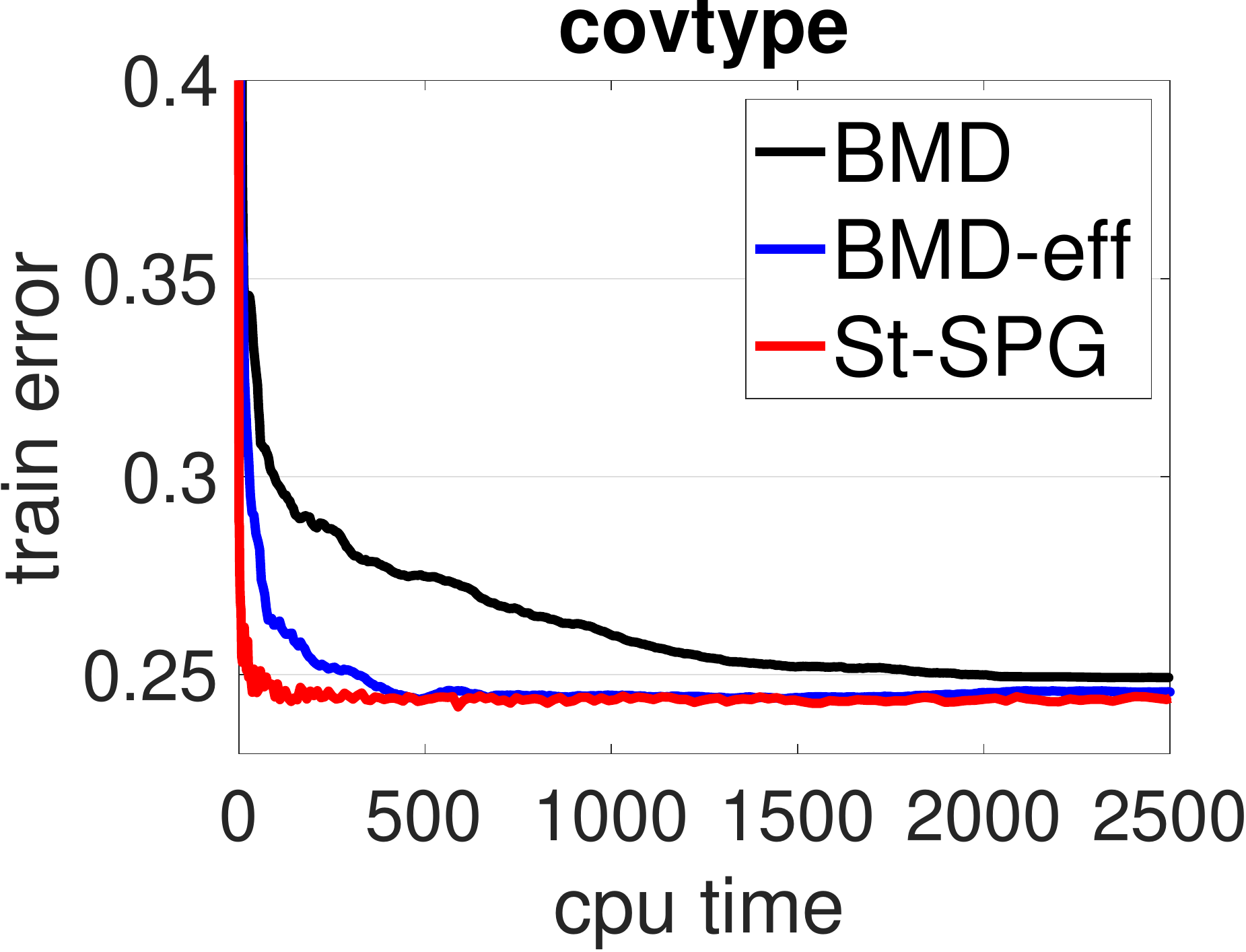}}
{\includegraphics[scale=.226]{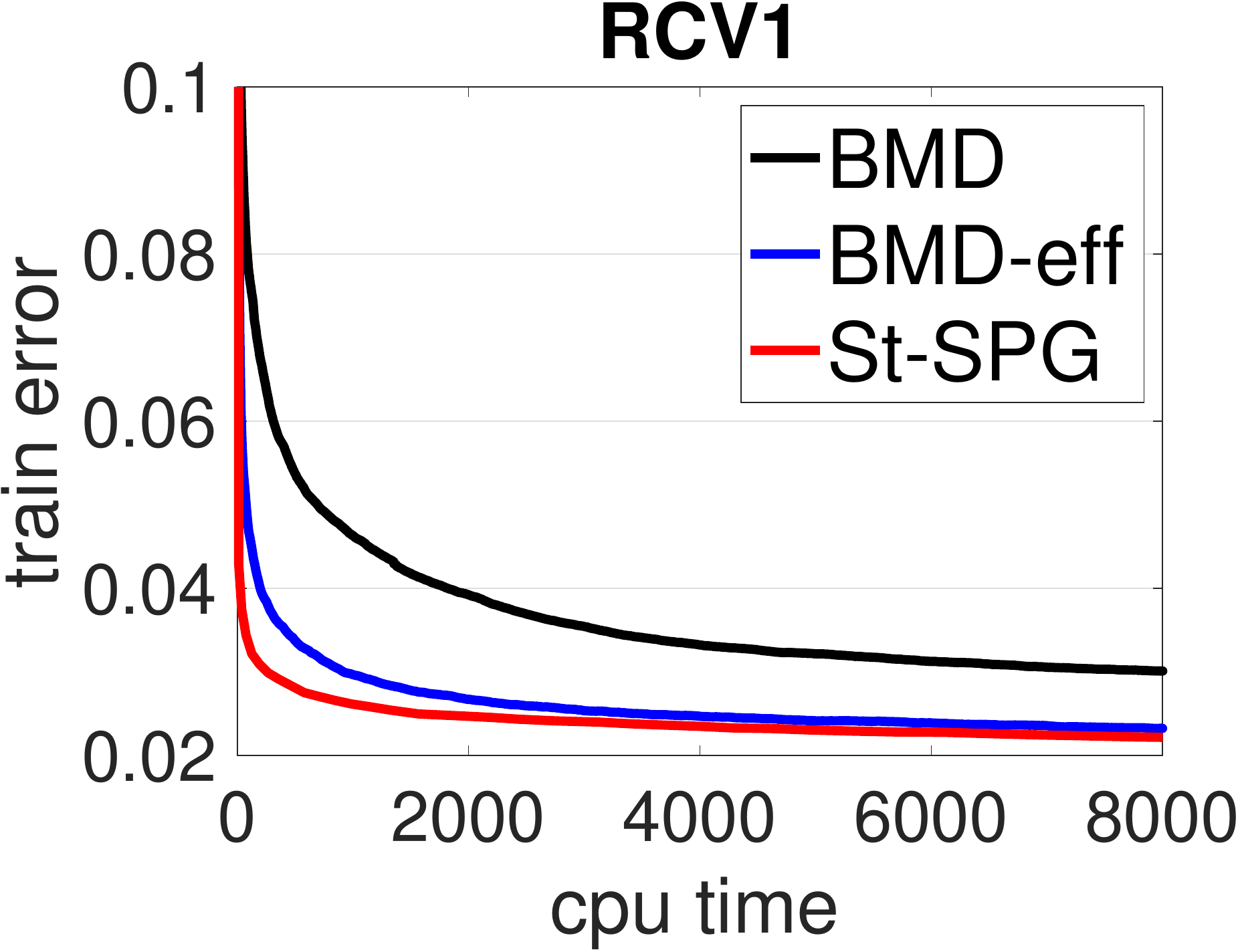}}
{\includegraphics[scale=.226]{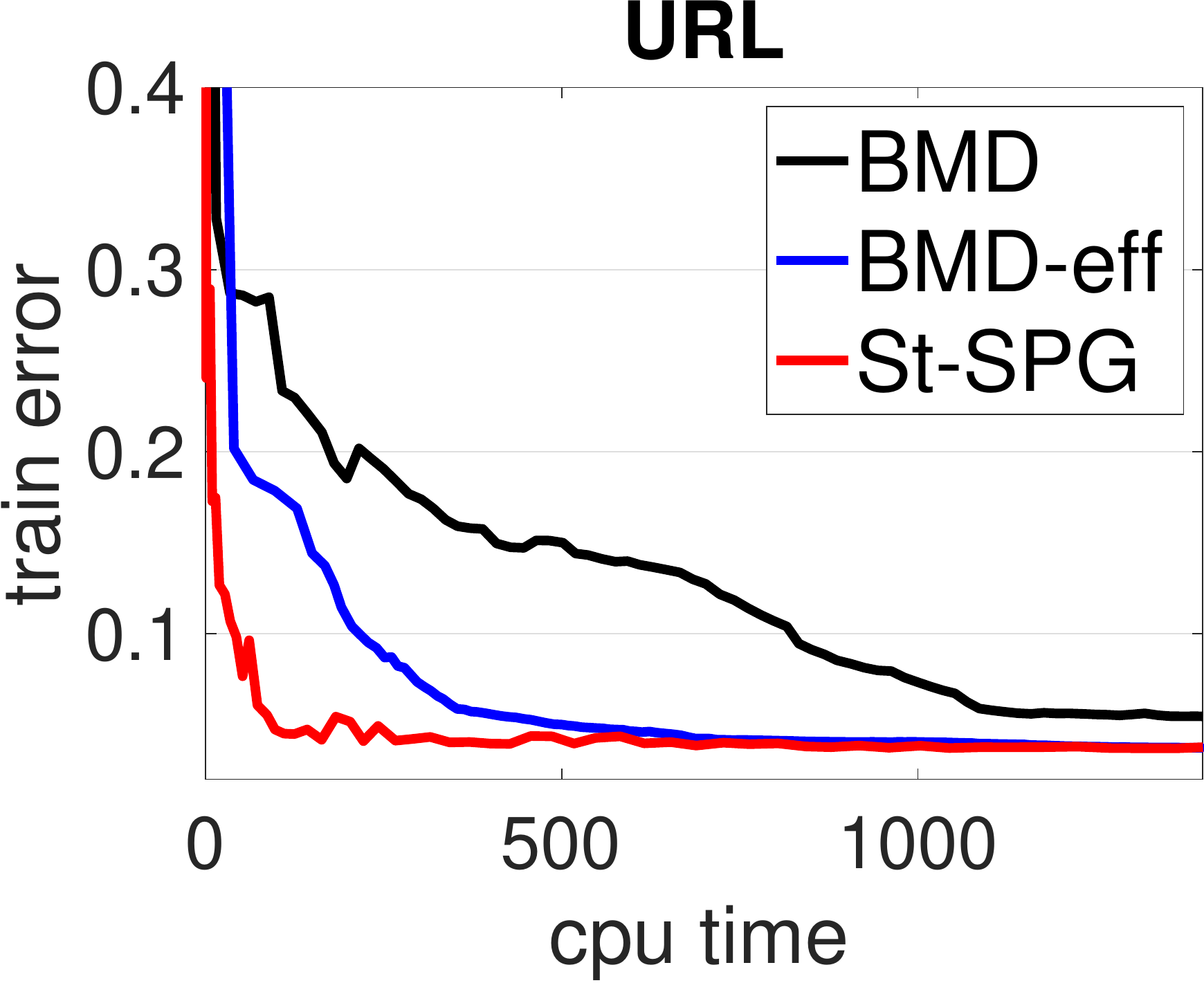}}
{\includegraphics[scale=.226]{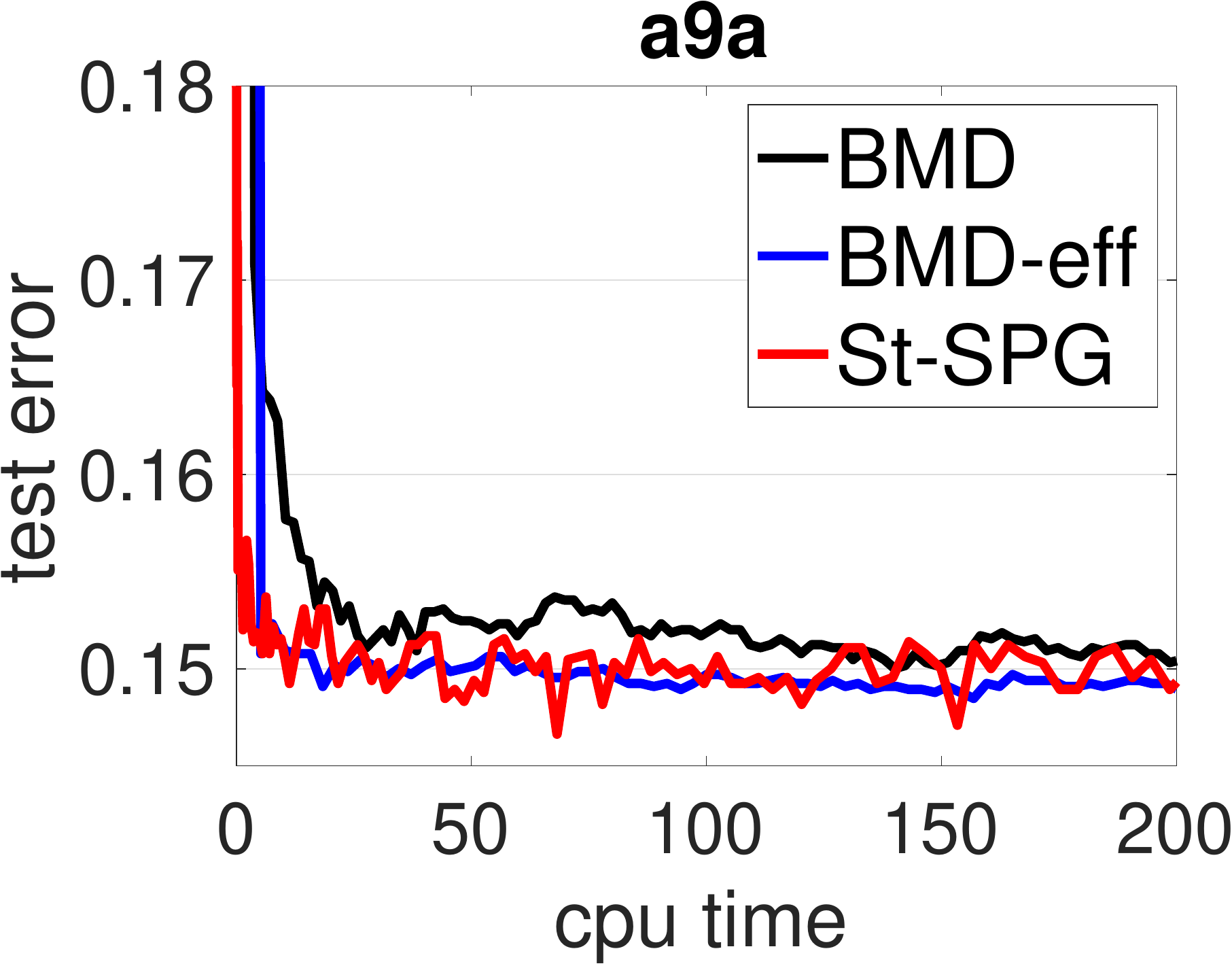}}
{\includegraphics[scale=.226]{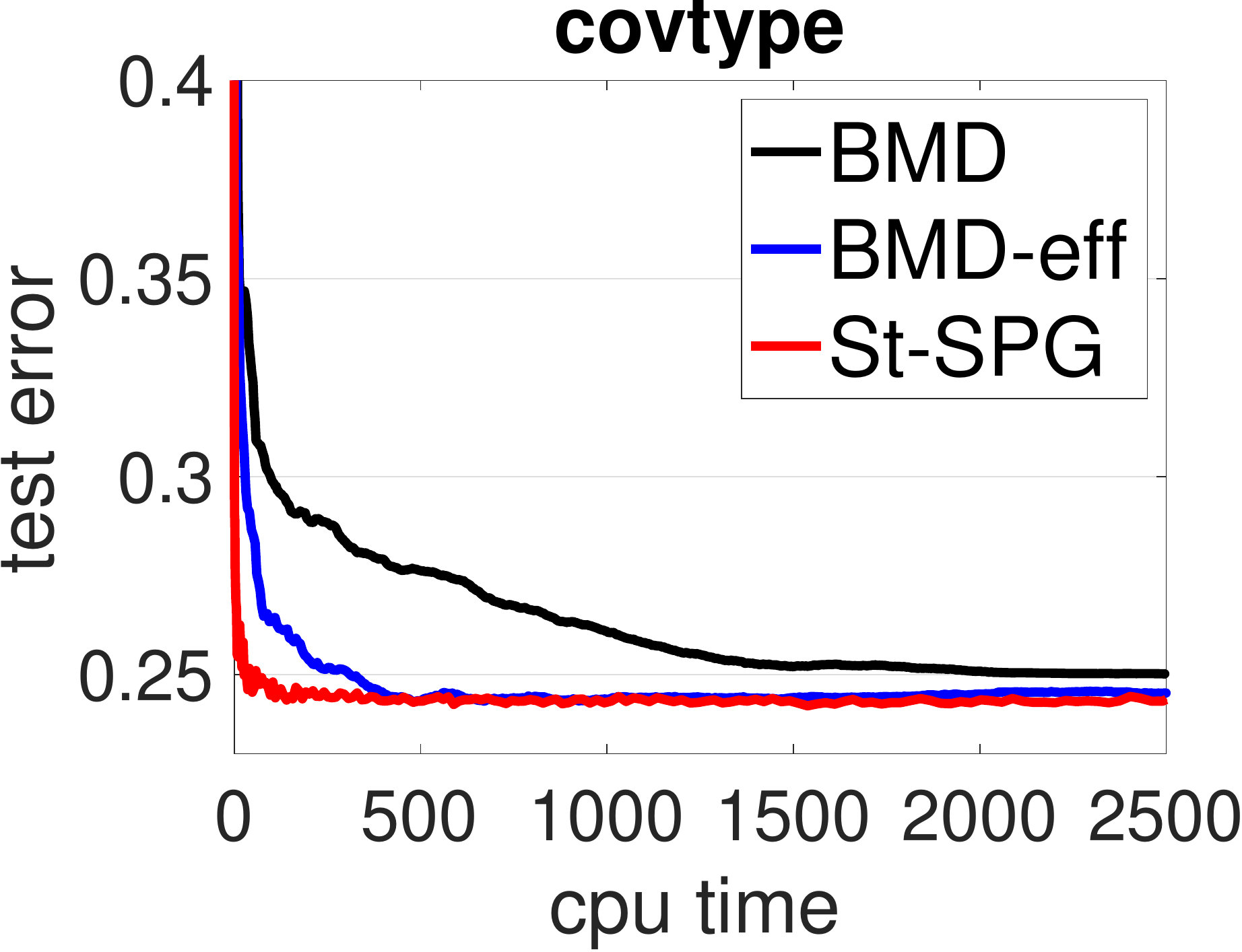}}
{\includegraphics[scale=.226]{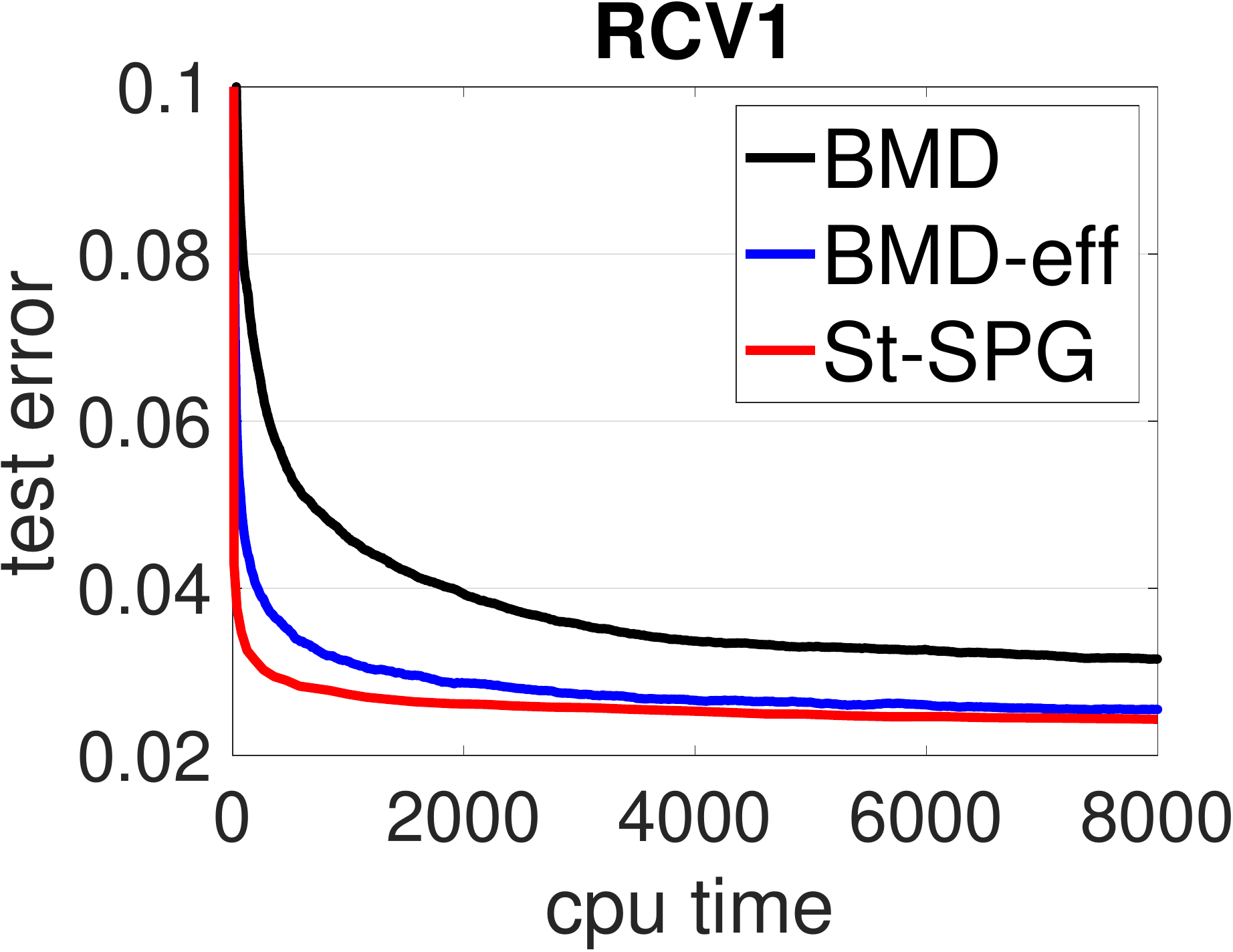}}
{\includegraphics[scale=.226]{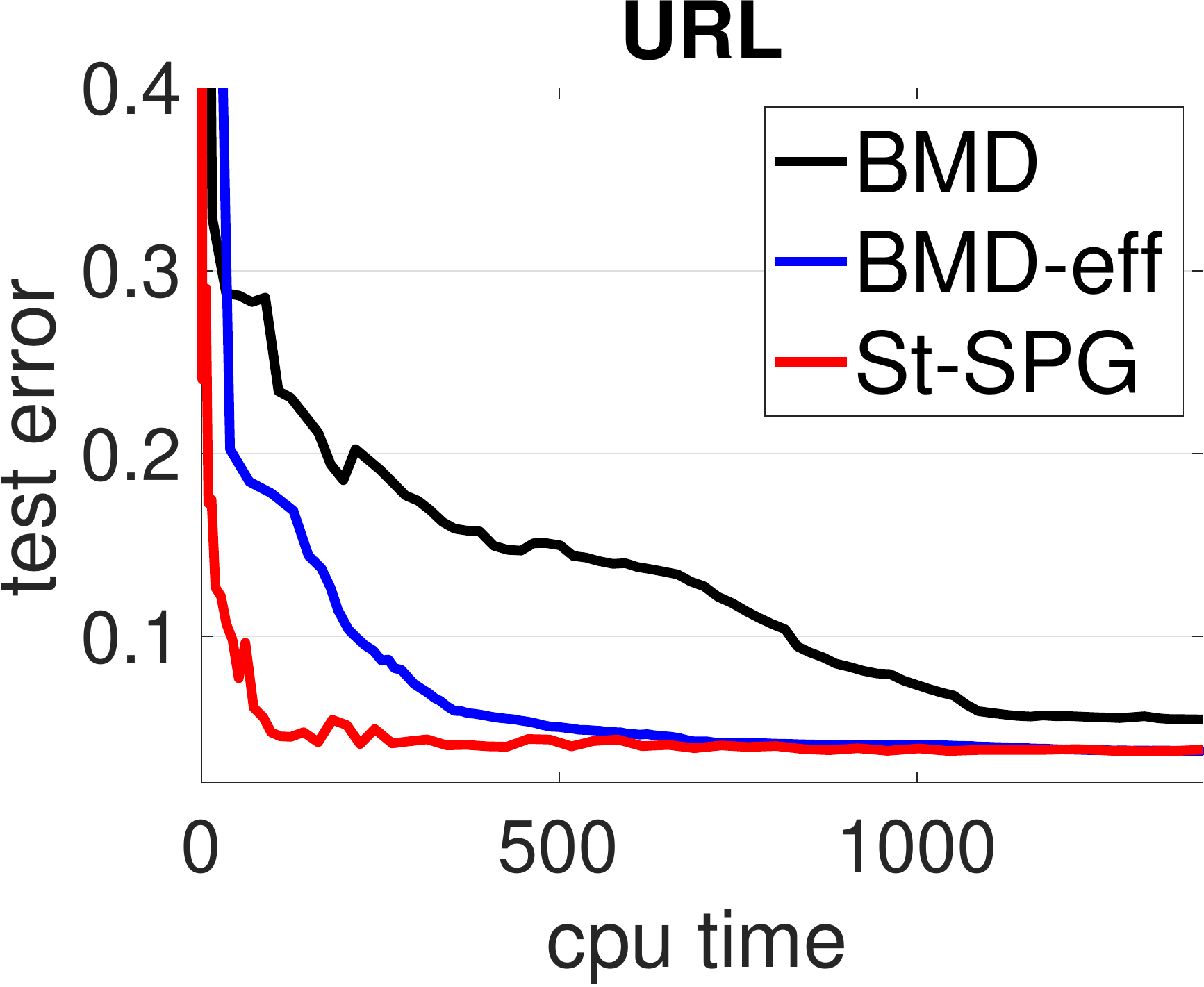}}
\caption{Results of variance-based regularization with (convex) logistic loss.}
\label{fig:dro_cputime}
\end{figure*}

\begin{figure*}[t]
\centering
{\includegraphics[scale=.23]{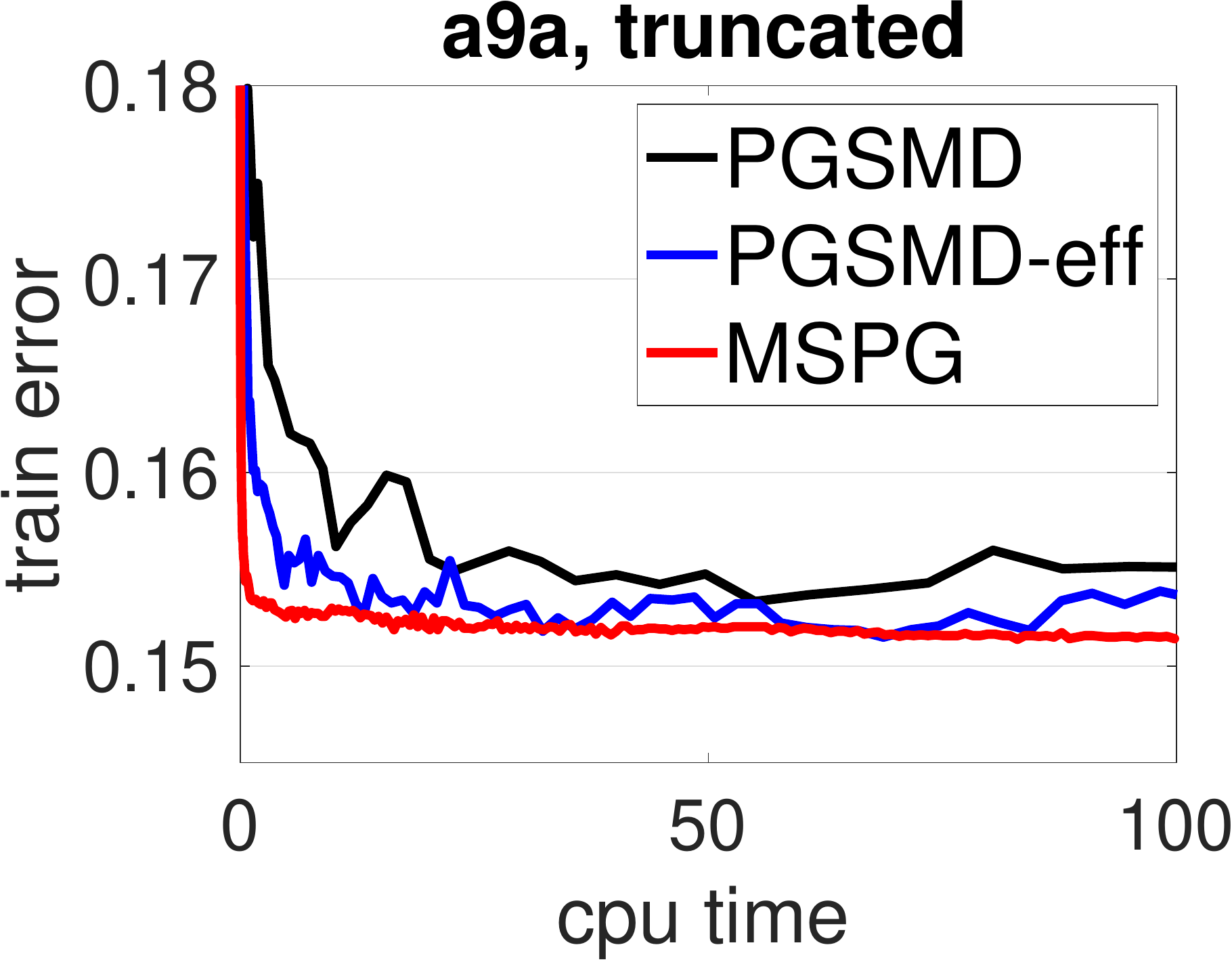}}
{\includegraphics[scale=.23]{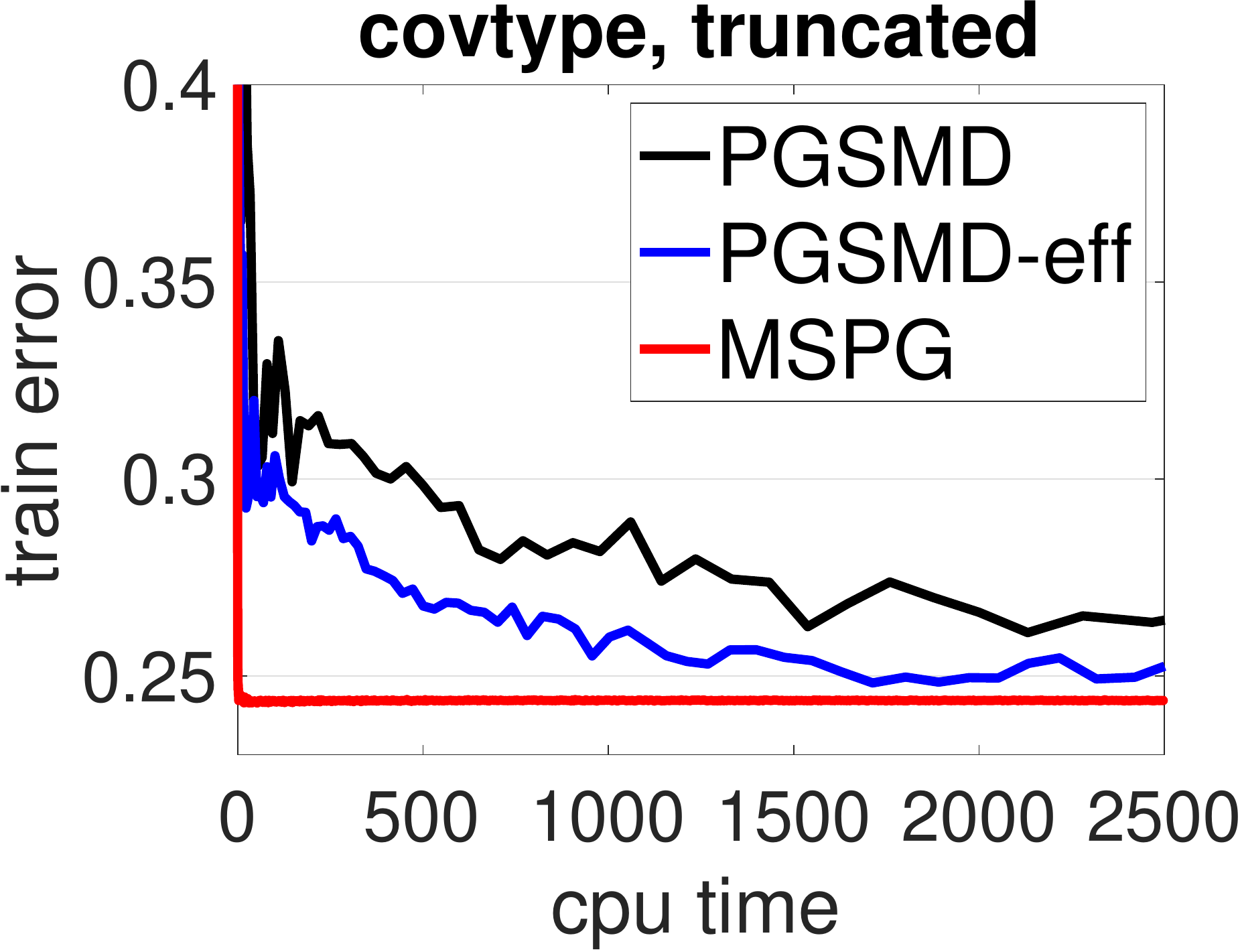}}
{\includegraphics[scale=.23]{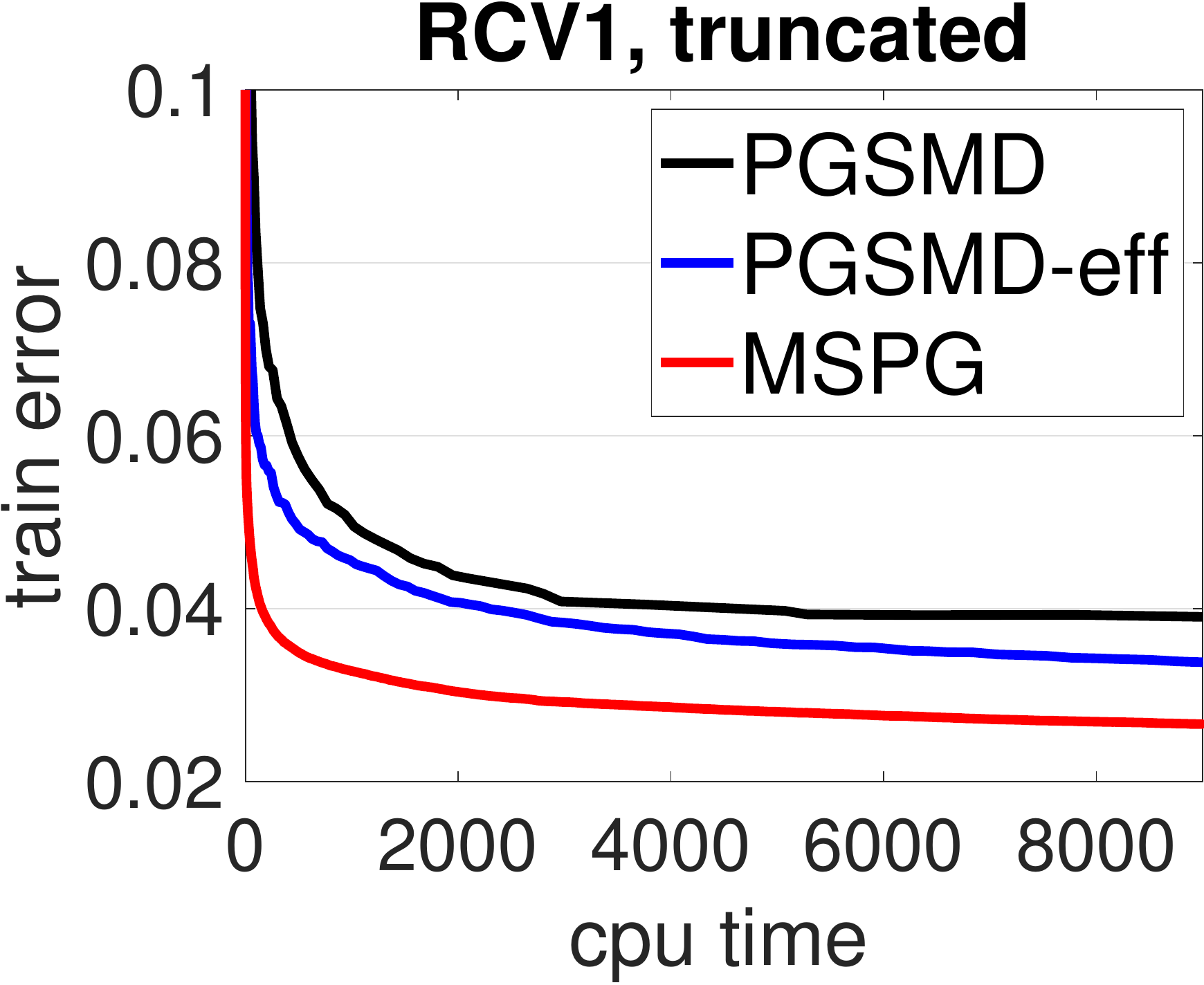}}
{\includegraphics[scale=.23]{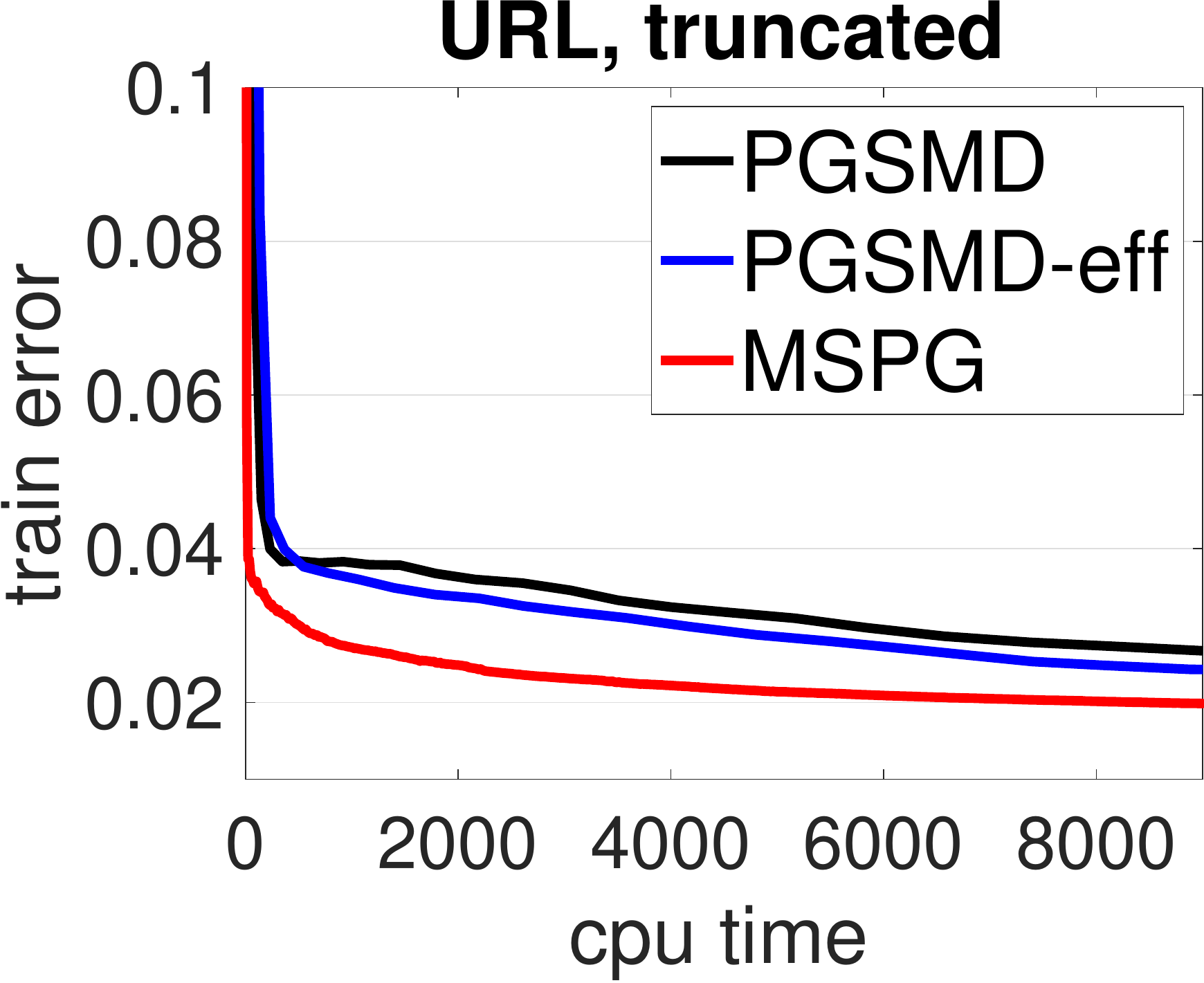}}
{\includegraphics[scale=.23]{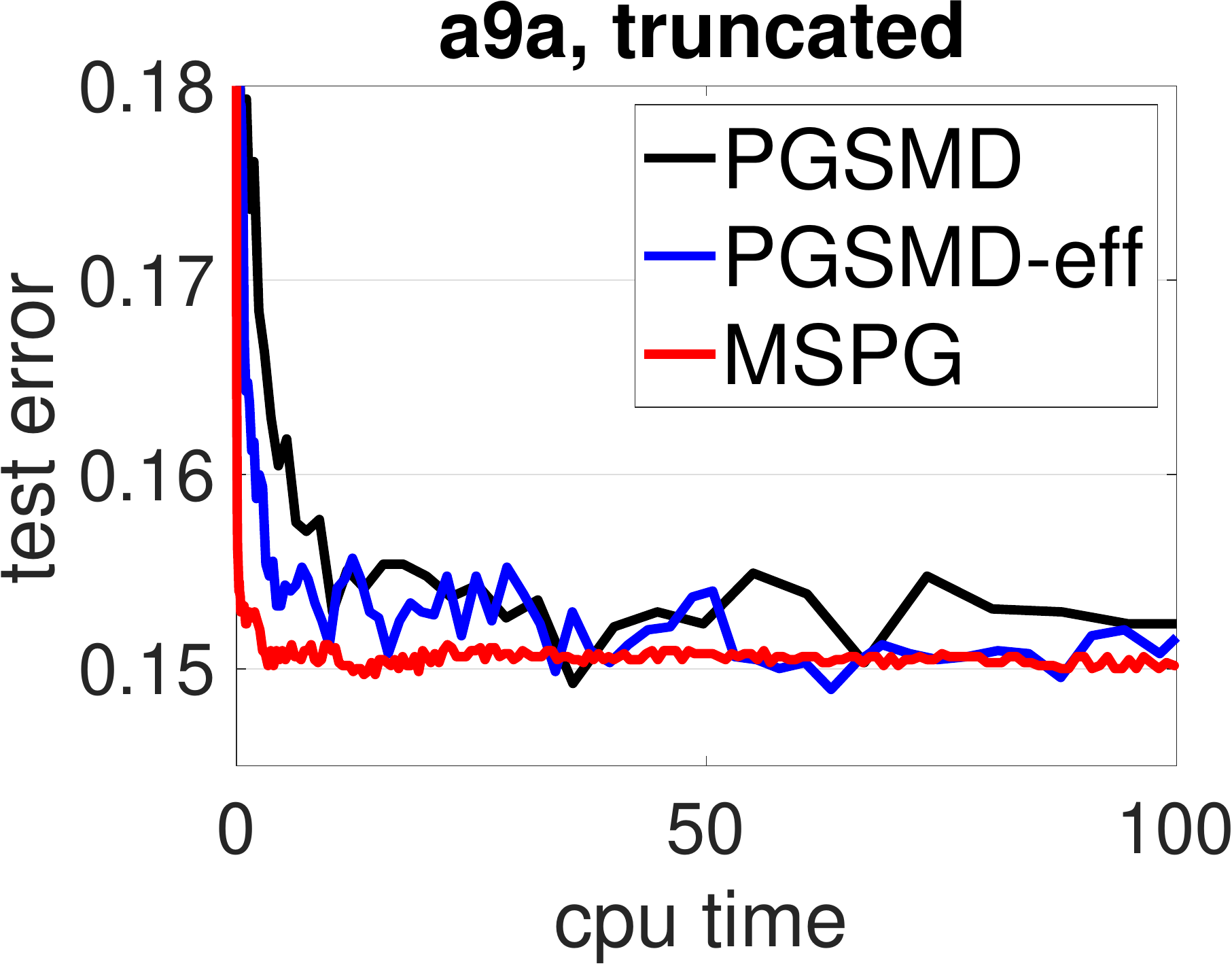}}
{\includegraphics[scale=.23]{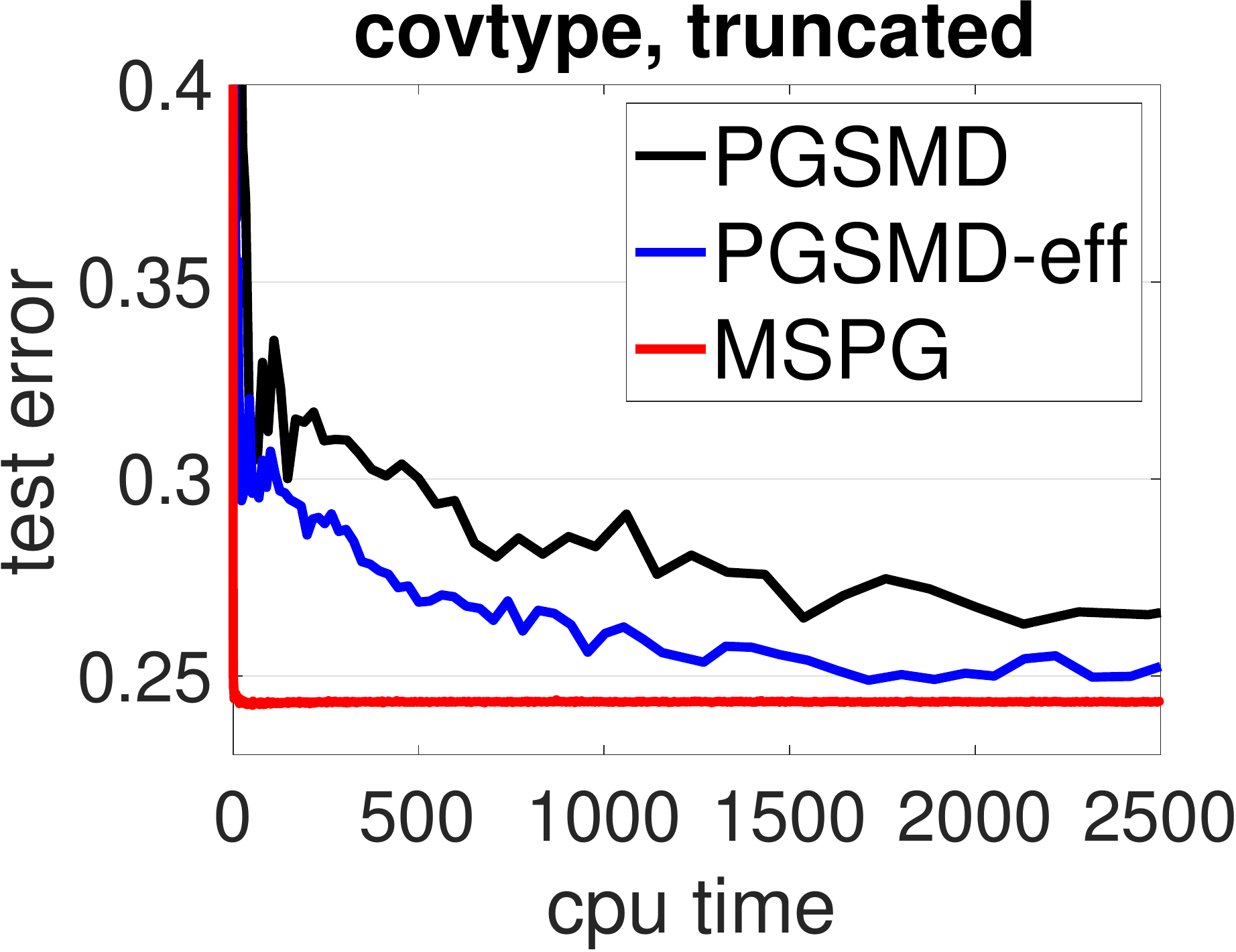}}
{\includegraphics[scale=.23]{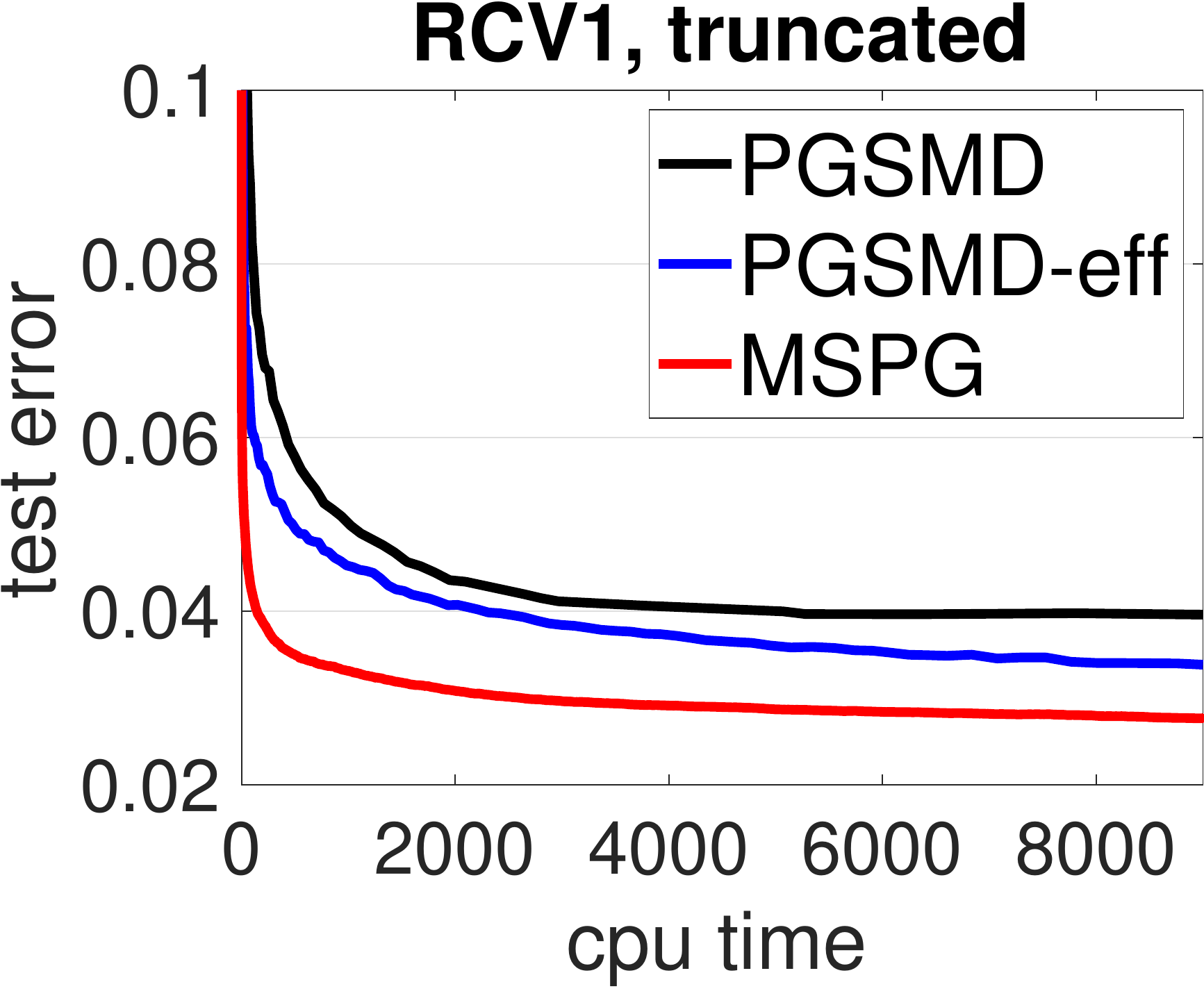}}
{\includegraphics[scale=.23]{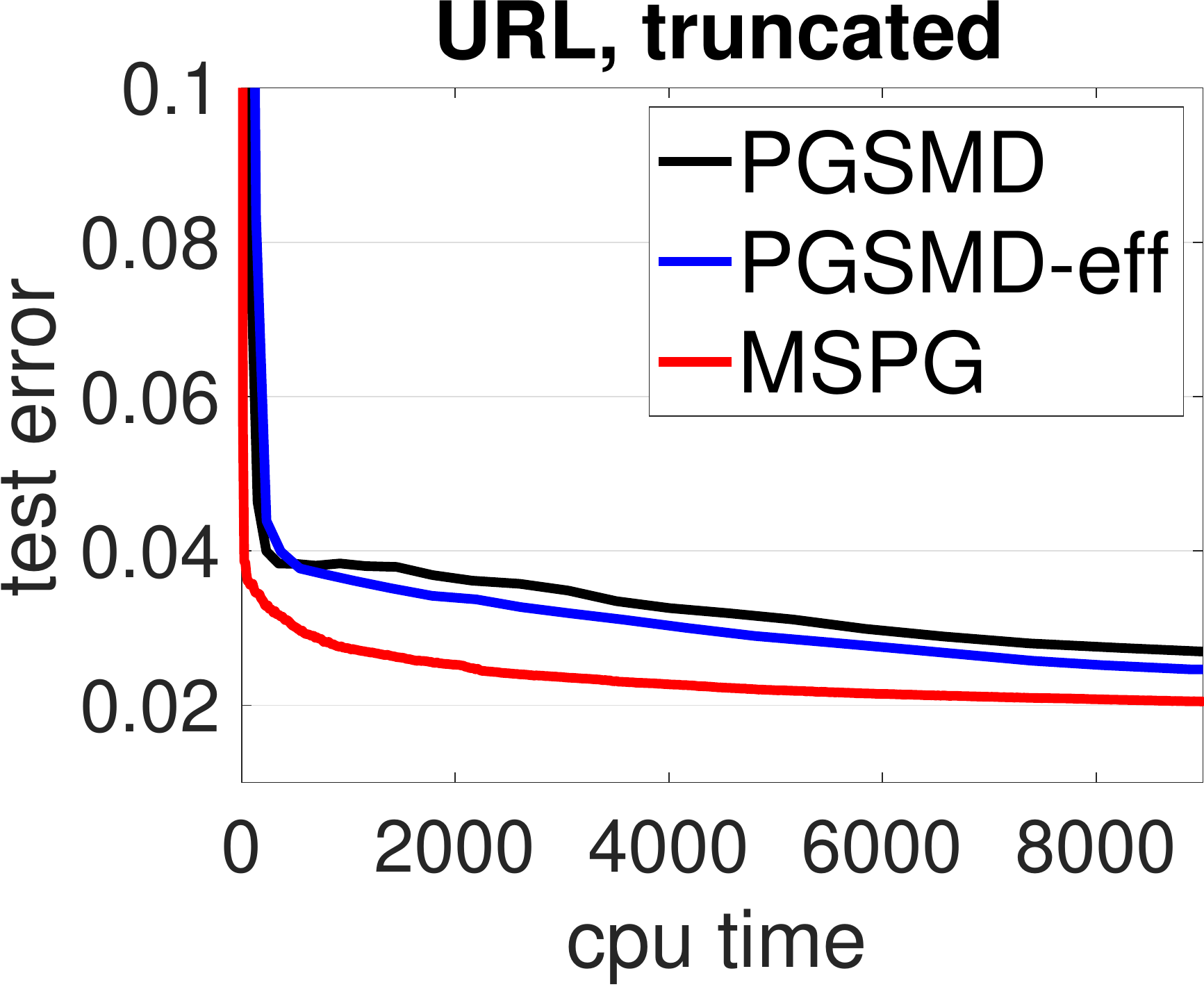}}
\caption{Results of variance-based regularization with for (non-convex) truncated logistic loss.}
\label{fig:dro_truncated_cputime}
\end{figure*}

We conduct experiments to verify the efficacy of the inf-projection formulation and proposed stochastic algorithms in comparison to the stochastic algorithms 
for solving min-max formulation~(\ref{eq:variance_regularized_robust}).
We perform two experiments on four datasets, i.e., a9a, RCV1, covtype and URL from the libsvm website, whose number of examples are $n=32561$,  $581012$, $697641$ and $2396130$, respectively (Table \ref{tab:data_stats}).
For each dataset, we randomly sample $80\%$ as training data and the rest as testing data.
We evaluate training error and testing error of our algorithms and baselines versus cpu time.

In the first experiment, we use (convex) logistic loss for $l(\theta, \z_i)$ in our inf-projection formulation (\ref{eq:dro_inf_projection_formulation}) and min-max formulation (\ref{eq:variance_regularized_robust}).
We compare our St-SPG with the stochastic algorithm Bandit Mirror Descend (BMD) proposed in \cite{DBLP:conf/nips/NamkoongD16}. 
We implement two versions of BMD, one using the standard mirror descent method to update the dual variable $P$ and the other (denoted by BMD-eff) exploiting binary search tree (BST) to update the $P$. To this end, it needs to use a modified constraint on $P$, i.e., $P \in \{ p \in \mathbb{R}_{+}^{n} | p_i \geq \delta/n, n^2 / 2 \| p - {\bf{1}}/n \|^2 \leq \rho \}$ (see Sec. 4 in \cite{DBLP:conf/nips/NamkoongD16}).
We tune hyper-parameters from a reasonable range, i.e., for St-SPG, $\lambda \in \{ 10^{-5:2} \}$, $\gamma, \mu \in \{ 10^{-3:3} \}$.
For BMD and BMD-eff, we tune step size $\eta_P \in \{ 10^{-8:-15} \}$ for updating $P$, step size  $\eta_\theta \in \{ 10^{-5:3} \}$ for updating $\theta$, $\rho \in \{ n \times 10^{-3:3} \}$ and fix $\delta = 10^{-5}$.
Training and testing errors against cpu time (s) of the three algorithms on four datasets are reported in Figure~\ref{fig:dro_cputime}.

In the second experiment, we use (non-convex) truncated logistic loss in (\ref{eq:dro_inf_projection_formulation}) and (\ref{eq:variance_regularized_robust}).
In particular, the truncated loss function is given by $\phi(l(\theta, \z_i)) = \alpha \log(1 + l(\theta, \z_i)/\alpha)$, where $l$ is logistic loss and we set $\alpha=\sqrt{10 n}$ as suggested in \cite{DBLP:journals/corr/abs-1805-07880}.
Since the loss is non-convex, we compare MSPG with proximally guided stochastic mirror descent (PGSMD) \cite{hassan18nonconvexmm} and its efficient variant (denoted by PGSMD-eff) for solving the min-max formulation that is non-convex and concave, where the efficient variant is implemented  with the same modified constraint on $P$ and BST as BMD-eff.
For MSPG, we tune $\lambda \in \{ 10^{-5:2} \}$, the step size parameter $c$ in Proposition~\ref{prop:1} from $\{ 10^{-5:2} \}$.
Hyper-parameters of PGSMD and PGSMD-eff including $\eta_P$, $\eta_\theta$, $\rho$ and $\delta$ are selected in the same range as in the first experiment.
The weak convexity parameter $\rho_{\text{wc}}$ are chosen from $\{ 10^{-5:5} \}$.
Training and testing errors against cpu time (s) of the three algorithms on four datasets are reported in Figure~\ref{fig:dro_truncated_cputime}.

We can observe two conclusions from the results of both experiments. First,  the training and testing errors from solving the inf-projection formulation (\ref{eq:dro_inf_projection_formulation}) converge to a close or even a lower level compared to that from solving the min-max formulation (\ref{eq:variance_regularized_robust}), which verifies the efficacy of the inf-projection formulation.
Second, the proposed stochastic algorithms have  significant improvement in the convergence time of training/testing errors, especially on large datasets, covtype, RCV1 and URL, which  can be verified by comparing convergence of training/testing errors against cpu time.

\section{Conclusion}
In this paper, we design and analyze stochastic optimization algorithms for a family of inf-projection minimization problems.
We show that the concerned inf-projection structure covers a variety of special cases, including DC functions and bi-convex functions as special cases (non-smooth functions in Section \ref{sec:practical_section}) and another family of inf-projection formulations (smooth functions in Section \ref{sec:smooth_section}). 
We develop stochastic optimization algorithms for those problems with theoretical guarantees of their first-order convergence for finding a (nearly) $\epsilon$-stationary solution at $O(1/\epsilon^{4/v})$.
To the best of our knowledge, this is the first work to provide comprehensive convergence analysis for stochastic optimization of non-convex inf-projection minimization problems.
Additionally, to verify the significance of our inf-projection formulation, we investigate an important machine learning problem, variance-based regularization, and compare our algorithms with baselines for min-max formulation (distributionally robust optimization).
Empirical results demonstrate the significance and effectiveness of our proposed algorithms.

\nocite{langley00}

\bibliographystyle{icml2020}
\bibliography{all,references}

\appendix

\newpage

\onecolumn

\appendix

\section{Proof in Section \ref{sec:smooth_section}}

\subsection{Proof of Lemma \ref{lemma:Holder_to_uniformly_convex}}

\begin{proof}
We prove the first part. The second part was proved in~\cite{Nesterov2015}. Recall that 
\begin{align}\label{eq:Holder_smooth_property}
f(x_1) - f(x_2) \leq \langle \nabla f(x_2), x_1 - x_2 \rangle + \frac{L}{1+v} \| x_1 - x_2 \|^{1+v} .
\end{align}

Define $\phi(y) = f(x + y) - f(x) - \langle \nabla f(x), y \rangle$.
By $(L, v)$-H\"older continuity of $\nabla f(x) $, one has
$\phi(y) \leq \frac{ L }{1 + v} \| y \|^{1 + v}_{2}$ due to (\ref{eq:Holder_smooth_property}).
Denote $\psi(y) = \frac{ L }{1 + v} \| y \|^{1 + v}_{2}$.

Given the definition of $\phi(y)$ and $\psi(y)$, we could derive their convex conjugates, denoted by $\phi^*(u)$ and $\psi^*(u)$.
For $\phi^*(u)$, one has
\begin{align*} 
\phi^*(u) = &
              \sup_{y} \langle y, u \rangle - \phi(y)
              \\
          = & 
              \sup_{y} \langle y, u \rangle - [ f(x + y) - f(x) - \langle \nabla f(x), y \rangle ]
              \\
          = & 
              \sup_{y} \langle y, u + \nabla f(x) \rangle - f(x + y) + f(x)
              \\
          \stackrel{ \text{\tiny{ \circled{1} }} }{=} & 
              \sup_{z} \langle z - x, u + \nabla f(x) \rangle - f(z) + f(x)
              \\
          = &
              \sup_{z} \langle z, u + \nabla f(x) \rangle - f(z) + f(x) - \langle x, u + \nabla f(x) \rangle 
              \\
          \stackrel{ \text{\tiny{ \circled{2} }} }{=} &
              f^*(u + \nabla f(x)) + f(x) - \langle x, \nabla f(x) \rangle - \langle x, u \rangle 
              \\
          \stackrel{ \text{\tiny{ \circled{3} }} }{=} &
              f^*(u + \nabla f(x)) - f^*(\nabla f(x)) - \langle x, u \rangle ,
\end{align*}
where {\tiny{ \circled{1} }} is due to letting $z = x + y$, {\tiny{ \circled{2} }} is due to the definition of convex conjugate and {\tiny{ \circled{3} }} is due to Fenchel-Young inequality (in this case, the equality holds), i.e, $f(x) + f^*( \nabla f(x) ) = \langle x, \nabla f(x) \rangle$.

For $\psi^*(u)$, one has
\begin{align*}
\psi^*(u) = &
              \sup_{y} \langle y, u \rangle - \frac{ L }{1 + v} \| y \|_{2}^{1 + v}
              \\
      \stackrel{ \text{\tiny{ \circled{1} }} }{=} & 
              \langle y^*(u), u \rangle - \frac{ L }{1 + v} \| y^*(u) \|_{2}^{1 + v}
              \\
          \stackrel{ \text{\tiny{ \circled{2} }} }{=} &
              \| y^*(u) \|_{2} \| u \|_{2} - \frac{ L }{1 + v} \| y^*(u) \|_{2}^{1 + v}
              \\
          \stackrel{ \text{\tiny{ \circled{3} }} }{=} &
              \Big( \frac{1}{L} \Big)^{ \frac{1}{v} } \| u \|_{2}^{1 + \frac{1}{v} }
              - \frac{ L }{ 1 + v } \Big(  \frac{1}{L} \| u \|_{2})  \Big)^{1 + \frac{1}{v} }
              \\
          = &
              \Big( 1 - \frac{1}{1 + v} \Big) \Big( \frac{1}{L} \Big)^{ \frac{1}{v} } \| u \|_{2}^{1 + \frac{1}{v} }   ,
\end{align*}
where \text{\tiny{ \circled{1} }} is due to letting $y^*(u) \in \arg\max_{y} \langle y, u \rangle = \frac{L}{1 + v} \| y \|_{2}^{1+v}$.
\text{\tiny{ \circled{2} }} is due to 
$u = L \| y^*(u) \|_{2}^{v - 1} \cdot y^*(u)$ and thus $\langle u, y^*(u) \rangle = \| u \|_{2} \cdot \| y^*(u) \|_{2}$.
\text{\tiny{ \circled{3} }} is due to 
$\| u \| \cdot \| y^*(u) \| 
= \langle u, y^*(u) \rangle 
= L \| y^*(u) \|_{2}^{v-1} \langle y^*(u), y^*(u) \rangle 
= L \| y^*(u) \|_{2}^{v+1}$.

Due to Lemma 19 of \cite{shalev2010equivalence}, if $\phi(y) \leq \psi(y) $, then one has $\phi^*(u) \geq \psi^*(u)$ and thus for all $u$ and $x$, 
\begin{align}\label{eq:prove_uniformly_convex1}
       f^*(u + \nabla f(x)) - f^*(\nabla f(x)) - \langle x, u \rangle
\geq 
       \Big( 1 - \frac{1}{1 + v} \Big) \Big( \frac{1}{L} \Big)^{ \frac{1}{v} } \| u \|_{2}^{1 + \frac{1}{v} }
\end{align}

Let $u'$ be any point in the relative interior of the domain of $f^*$.
Then we need to prove that if $x \in \partial f^*(u')$, then $u' = \nabla f(x)$.
By Fenchel-Young inequality, one has
$
\langle x, u' \rangle = f(x) + f^*(u')  
$
and
$
\langle x, \nabla f(x) \rangle = f(x) + f^*(\nabla f(x))  .
$
By (\ref{eq:prove_uniformly_convex1}), 
\begin{align*}
      0
=    &
       f(x) - f(x)
       \\
=    & 
       \langle x, \nabla f(x) \rangle - f^*(\nabla f(x)) - ( \langle x , u' \rangle - f^*(u') )
       \\
=    & 
       f^*(u') - f^*(\nabla f(x) )  - \langle x, u' - \nabla f(x) \rangle
       \\
\geq &
       \Big( 1 - \frac{1}{1 + v} \Big) \Big( \frac{1}{L} \Big)^{\frac{1}{v}} \| u' - \nabla f(x) \|_{2}^{1 + \frac{1}{v} }   ,
\end{align*}
which implies that $u' = \nabla f(x)$.
Thus,
\begin{align*}
       f^*(u + u') - f^*(u') - \langle \partial f^*(u'), u \rangle
\geq 
       \frac{1}{2} \cdot \frac{2v}{1 + v} \Big( \frac{1}{L
       } \Big)^{ \frac{1}{v} } \| u \|_{2}^{1 + \frac{1}{v} }  
\end{align*}
implies $f^*$ is $(\varrho, p)$-uniformly convex with
$\varrho = \frac{2v}{1+v} \frac{1}{L^{\frac{1}{v}}}$
and
$p = 1 + \frac{1}{v}$.
\end{proof}

\subsection{Proof of Lemma \ref{lemma:joint_L_smooth1}}

\begin{proof}
First consider
\begin{align}\label{eq:joint_smooth_x_part}
     & \| \nabla_{x} f_0(x, y) - \nabla_{x} f_0(x', y') \|_{2}^{2}
       \nonumber\\
=    & 
       \| \nabla g(x) - \nabla g(x') - \nabla\ell(x)^{\top} y + \nabla\ell(x')^{\top} y' \|_{2}^{2}
       \nonumber\\
\leq &
       2 \| \nabla g(x) - \nabla g(x') \|_{2}^{2} 
       + 2 \|  \nabla\ell(x)^{\top} y - \nabla\ell(x')^{\top} y + \nabla\ell(x')^{\top} y - \nabla\ell(x')^{\top} y'  \|_{2}^{2}   
       \nonumber\\
\leq &
       2 L_{g}^{2} \| x - x' \|_{2}^{2} 
       + 4 D_{y}^{2} \| \nabla\ell(x) - \nabla\ell(x') \|_{2}^{2} 
       + 4 G_{\ell}^{2} \| y - y' \|_{2}^{2}
       \nonumber\\
\leq &
       2 L_{g}^{2} \| x - x' \|_{2}^{2} 
       + 4 L_{\ell}^{2} D_{y}^{2} \| x - x' \|_{2}^{2} 
       + 4 G_{\ell}^{2} \| y - y' \|_{2}^{2}
\end{align}

Then consider
\begin{align}\label{eq:joint_smooth_y_part}
     & \| \nabla_{y} f_0(x, y) - \nabla_{y} f_0(x', y') \|_{2}^{2}
       \nonumber\\
=    & 
       \| \ell(x)  -  \ell(x') \|_{2}^{2}
\leq 
        G_{\ell}^{2} \| x - x' \|_{2}^{2}   .
\end{align}

Combining the above two inequalities (\ref{eq:joint_smooth_x_part}) and (\ref{eq:joint_smooth_y_part}), one has
\begin{align*}
     & \| \nabla_{x} f(x, y) - \nabla_{x} f(x', y') \|_{2}^{2}
    + 
     \| \nabla_{y} f(x, y) - \nabla_{y} f(x', y') \|_{2}^{2}
     \\
\leq &
       (2 L_{g}^{2} + 4L_{\ell}^{2} D_{y}^{2} + G_{\ell}^{2} ) \| x - x' \|_{2}^{2} 
       + 4 G_{\ell}^{2}  \| y - y' \|_{2}^{2}  
       \\
\leq & 
       L^{2} ( \| x - x' \|_{2}^{2} + \| y - y' \|_{2}^{2} )   ,
\end{align*}
where $L = \sqrt{ \Big( \max(2L_{g}^{2} + 4 L_{\ell}^{2} D_{y}^{2} + G_{\ell}^{2} ,  4G_{\ell}^{2} \Big) }$.
\end{proof}

\subsection{Proof of Proposition~\ref{prop:1}}
\begin{proof}
This analysis is borrowed from the proof of Theorem 2 in~\cite{DBLP:journals/corr/abs/1902.07672}. For completeness, we include it here.
Let $w=(x,y)$, $\nabla_x f^{(t)}_0= \nabla_{x} f_0(x_{t}, y_{t})$, $\nabla_y f^{(t)}_0=\nabla_{y} f_0(x_{t}, y_{t})$, $\nabla f^{(t)}_0=(\nabla_x f^{(t)}_0,\nabla_y f^{(t)}_0)$ and $\widetilde\nabla f^{(t)}_0=(\widetilde\nabla_x f^{(t)}_0,\widetilde\nabla_y f^{(t)}_0)$. By the udpate of $x_{t+1}=\Pi_{X}[x_{t} - \eta \widetilde\nabla_{x} f^{(t)}_0]$, we know
\begin{align*}
x_{t+1}
=  & \arg\min_{x\in\bbR^d}\left\{ I_{X}(x)+\langle \widetilde\nabla_{x} f^{(t)}_0, x-x_t\rangle + \frac{1}{2\eta }\|x- x_t\|^2\right\}.
\end{align*}
 and
\begin{align}\label{prop1:ineq:x}
 \langle \widetilde\nabla_{x} f^{(t)}_0, x_{t+1}-x_t\rangle + \frac{1}{2\eta }\|x_{t+1}- x_t\|^2 \leq 0.
\end{align}
Similarly, by the update of $y_{t+1} =P_{\eta h} [y_{t} - \eta\widetilde\nabla_{y} f^{(t)}_0]$, we know
\begin{align*}
y_{t+1}
=  \arg\min_{y\in\text{dom}(h)}\left\{ h(y) +\langle \widetilde\nabla_{y} f^{(t)}_0, y-y_t\rangle + \frac{1}{2\eta }\|y- y_t\|^2\right\},
\end{align*}
and
\begin{align}\label{prop1:ineq:y}
h(y_{t+1}) +\langle \widetilde\nabla_{y} f^{(t)}_0, y_{t+1}-y_t\rangle + \frac{1}{2\eta }\|y_{t+1}- y_t\|^2 \leq h(y_t).
\end{align}
Using the inequalities (\ref{prop1:ineq:x}) and (\ref{prop1:ineq:y}), and the fact that $w=(x,y)$, we get
\begin{align}\label{prop1:ineq:z}
 h(y_{t+1})+\langle \widetilde\nabla f^{(t)}_0, w_{t+1}-w_t\rangle + \frac{1}{2\eta }\|w_{t+1}- w_t\|^2 \leq h(y_t).
\end{align}
We know from Lemma~\ref{lemma:joint_L_smooth1} that $f_0(w)$ is $L$-smooth, thus
\begin{align}\label{prop1:ineq:smooth}
 f_0(w_{t+1}) \leq f_0(w_t) +\langle \nabla f_0^{(t)}, w_{t+1}-w_t\rangle + \frac{L}{2}\|w_{t+1}- w_t\|^2. 
\end{align}
Combining the inequalities (\ref{prop1:ineq:z}) and (\ref{prop1:ineq:smooth}) and using the fact that $f(w) = f_0(w) + h(y)$we have
\begin{align}\label{prop1:ineq1}
  \frac{1-\eta L}{2\eta}\|w_{t+1}- w_t\|^2 \leq f(w_t) -f(w_{t+1}) +\langle \nabla f^{(t)}_0 - \widetilde\nabla f^{(t)}_0, w_{t+1}-w_t\rangle.
 \end{align}
 Applying Young's inequality $\langle a, b\rangle \leq \frac{1}{2L}\|a\|^2 + \frac{L}{2}\|b\|^2$ to the last inequality of (\ref{prop1:ineq1}), we then have
\begin{align}\label{prop1:ineq2}
  \frac{1-2\eta L}{2\eta}\|w_{t+1}- w_t\|^2 \leq f(w_t) -f(w_{t+1}) +\frac{1}{2L}\|\nabla f^{(t)}_0 - \widetilde\nabla f^{(t)}_0\|^2.
 \end{align}
 Summing (\ref{prop1:ineq2}) across $t=0,\dots, T-1$, we have
\begin{align}\label{prop1:ineq3}
\nonumber \frac{1-2\eta L}{2\eta}\sum_{t=0}^{T-1}\|w_{t+1}- w_t\|^2\leq&  f(w_0) -  f(w_{T}) + \frac{1}{2L}\sum_{t=0}^{T-1}\|\nabla f^{(t)}_0 - \widetilde\nabla f^{(t)}_0\|^2\\
   \leq & M + \frac{1}{2L}\sum_{t=0}^{T-1}\|\nabla f^{(t)}_0 - \widetilde\nabla f^{(t)}_0\|^2,
\end{align}
where the last inequality uses the Assumption~\ref{assumption:loss_function} (v).

Next, by Exercise 8.8 and Theorem 10.1 of \citep{RockWets98}, we know from the updates of $x_{t+1}$ and $y_{t+1}$ that
\begin{align*}
-\widetilde\nabla_{x} f^{(t)}_0 - \frac{1}{\eta }(x_{t+1}- x_t) \in \hat \partial I_{X}(x_{t+1}),\\
  -\widetilde\nabla_{y} f^{(t)}_0 - \frac{1}{\eta }(y_{t+1}- y_t) \in \hat \partial h(y_{t+1}),
\end{align*}
and thus
\begin{align}\label{prop1:stationary:z}
 \nabla f^{(t+1)}_0-\widetilde\nabla f^{(t)}_0 - \frac{w_{t+1}- w_t}{\eta } \in  \nabla f^{(t+1)}_0+ (\hat \partial h(y_{t+1}), \hat \partial I_{X}(x_{t+1}))= \hat\partial f(w_{t+1}).
\end{align}
Multiplying $\frac{2}{\eta}$ on both sides of (\ref{prop1:ineq1}) we get
\begin{align}\label{prop1:ineq4}
  \nonumber &\frac{2}{\eta}\langle\widetilde\nabla f^{(t)}_0- \nabla f^{(t+1)}_0 , w_{t+1}-w_t\rangle + \frac{1-\eta L}{\eta^2}\|w_{t+1}- w_t\|^2 \\
  \leq& \frac{2(f(w_t) -f(w_{t+1}))}{\eta} +\frac{2}{\eta}\langle \nabla f^{(t)}_0 -\nabla f^{(t+1)}_0, w_{t+1}-w_t\rangle.
 \end{align}
 By the fact that $\frac{2}{\eta}\langle\widetilde\nabla f^{(t)}_0- \nabla f^{(t+1)}_0 , w_{t+1}-w_t\rangle  = \| \widetilde\nabla f^{(t)}_0- \nabla f^{(t+1)}_0 + \frac{w_{t+1}-w_t}{\eta}\|^2 - \| \widetilde\nabla f^{(t)}_0- \nabla f^{(t+1)}_0 \|^2 -\frac{1}{\eta^2} \| w_{t+1}-w_t\|^2$, then
\begin{align*}
&\| \widetilde\nabla f^{(t)}_0- \nabla f^{(t+1)}_0 + \frac{w_{t+1}-w_t}{\eta}\|^2  \\
  \leq& \| \widetilde\nabla f^{(t)}_0- \nabla f^{(t+1)}_0 \|^2 + \frac{L}{\eta}\|w_{t+1}- w_t\|^2+\frac{2(f(w_t) -f(w_{t+1}))}{\eta} +\frac{2}{\eta}\langle \nabla f^{(t)}_0 -\nabla f^{(t+1)}_0, w_{t+1}-w_t\rangle\\
 \leq& \| \widetilde\nabla f^{(t)}_0- \nabla f^{(t+1)}_0 \|^2 + \frac{L}{\eta}\|w_{t+1}- w_t\|^2+\frac{2(f(w_t) -f(w_{t+1}))}{\eta} +\frac{2L}{\eta}\|w_{t+1}-w_t\|^2\\
  \leq& 2\| \widetilde\nabla f^{(t)}_0- \nabla f^{(t)}_0 \|^2 + 2\| \nabla f^{(t)}_0- \nabla f^{(t+1)}_0 \|^2 + \frac{3L}{\eta}\|w_{t+1}- w_t\|^2+\frac{2(f(w_t) -f(w_{t+1}))}{\eta}\\
    \leq& 2\| \widetilde\nabla f^{(t)}_0- \nabla f^{(t)}_0 \|^2 + (2L^2+\frac{3L}{\eta})\|w_{t+1}- w_t\|^2+\frac{2(f(w_t) -f(w_{t+1}))}{\eta},
\end{align*}
 where the second inequality is due to Cauchy-Schwartz inequality and the smoothness of $f(w)$; the third inequality is due to Young's inequality; and the last inequality is due to the smoothness of $f(w)$. 
 Summing  above inquality across $t=0,\dots, T-1$, we have
\begin{align*}
&\sum_{t=0}^{T-1}\| \widetilde\nabla f^{(t)}_0- \nabla f^{(t+1)}_0 + \frac{w_{t+1}-w_t}{\eta}\|^2  \\
    \leq& 2\sum_{t=0}^{T-1}\| \widetilde\nabla f^{(t)}_0- \nabla f^{(t)}_0 \|^2 + (2L^2+\frac{3L}{\eta})\sum_{t=0}^{T-1}\|w_{t+1}- w_t\|^2+\frac{2(f(w_0) -f(w_{T+1}))}{\eta}\\
 \leq& 2\sum_{t=0}^{T-1}\| \widetilde\nabla f^{(t)}_0- \nabla f^{(t)}_0 \|^2 + \frac{2}{\eta^2}\sum_{t=0}^{T-1}\|w_{t+1}- w_t\|^2+\frac{2M}{\eta},
 \end{align*}
where the last inequality uses the Assumption~\ref{assumption:loss_function} (v). 
Combining above inequality with (\ref{prop1:ineq3}) and  (\ref{prop1:stationary:z}) we obtain
\begin{align*}
&\E_R[\text{dist}(0,\hat \partial f(w_{R}))^2] \\
\leq & \frac{1}{T} \sum_{t=0}^{T-1}\E[\|\nabla f^{(t+1)}_0-\widetilde\nabla f^{(t)}_0 - \frac{(x_{t+1}- x_t, y_{t+1}- y_t) }{\eta }\|^2]\\
\leq & \frac{2}{T} \sum_{t=0}^{T-1}\E[\| \widetilde\nabla f^{(t)}_0- \nabla f^{(t)}_0 \|^2]+\frac{2M}{\eta T}
+ \frac{2}{\eta(1-2\eta L)}\left(2M + \frac{1}{L}\sum_{t=0}^{T-1}\E[\| \widetilde\nabla f^{(t)}_0- \nabla f^{(t)}_0 \|^2]\right)\\
= & \frac{2c(1 -2c)+2}{c(1-2c)}\frac{1}{T}\sum_{t=0}^{T-1}\E[\| \widetilde\nabla f^{(t)}_0- \nabla f^{(t)}_0 \|^2]+ \frac{6-4c}{1-2c}\frac{M}{\eta T}\\
\leq &  \frac{2c(1 -2c)+2}{c(1-2c)}\frac{1}{T}\sum_{t=0}^{T-1}\frac{\sigma_0^2}{b(t+1)}+ \frac{6-4c}{1-2c}\frac{M}{\eta T}\\
\leq &  \frac{2c(1 -2c)+2}{c(1-2c)}\frac{\sigma_0^2(\log(T)+1)}{bT}+ \frac{6-4c}{1-2c}\frac{M}{\eta T},
\end{align*}
where $0<c<\frac{1}{2}$, the last second inequalit is due to the bounded variance of stochastic gradient and the last inequality uses the fact that $\sum_{t=1}^{T} \frac{1}{t} \leq \log(T)+1$.
\end{proof}

\subsection{Proof of Lemma \ref{lemma:asymptotic_one_loop}}

\begin{proof} 
First, we derive $\nabla F(\tilde{x})$ for any $\tilde{x}$ as follows
\begin{align*}
       \nabla F(\tilde{x})
\stackrel{\text{\tiny{\circled{1}}}}{=}    
     &
       \nabla g(\tilde{x}) - \nabla\ell(\tilde{x})^{\top} y^*(\tilde{x}) 
       \\
=    &
       \nabla g(\tilde{x}) - \nabla\ell(\tilde{x})^{\top} \tilde{y}
       + \nabla\ell(\tilde{x})^{\top} ( \tilde{y} - y^*(\tilde{x}) ) 
       \\
=    &
       \nabla_{x} f(\tilde{x}, \tilde{y}) 
       + \nabla\ell(\tilde{x})^{\top} ( \tilde{y} - y^*(\tilde{x}) )  ,
\end{align*}
where $\nabla\ell(x)$ is the Jacobian matrix of $\ell$ at $x$,
and $y^*(x) = \arg\min_{y \in \dom(h)} h(y) - \langle y, \ell(x) \rangle = \arg\min_{y\in \dom(h)} f(x, y)$.
Here $y^*(x)$ is unique given $x$, since uniform convexity ensures the unique solution ($\nabla h^*$ is H\"older continuous so that $h$ is uniformly convex).
Equality \text{\tiny{\circled{1}}} above is due to Theorem 10.58 of \cite{rockafellar2009variational} and unique $y^*(\tilde{x})$.

Then by triangle inequality and Cauchy-Schwarz inequality, one has
\begin{align*}
       \| \nabla F(\tilde{x}) \|_{2}
\leq & 
       \| \nabla_{x} f(\tilde{x}, \tilde{y})  \|_{2}
       + \| \nabla\ell(\tilde{x})^{\top} ( \tilde{y} - y^*(\tilde{x}) ) \|_{2}
       \\
\leq & 
       \| \nabla_{x} f(\tilde{x}, \tilde{y})  \|_{2}
       + \| \nabla\ell(\tilde{x}) \|_{2} \cdot \| ( \tilde{y} - y^*(\tilde{x}) ) \|_{2}
       \\    
\leq & 
       \| \nabla_{x} f(\tilde{x}, \tilde{y})  \|_{2}
       + G_{\ell} \Big( \frac{1}{\varrho} \| \partial_{y} f(\tilde{x}, \tilde{y}) \|_{2} \Big)^{\frac{1}{p-1}}
       \\        
=    & 
       \| \nabla_{x} f(\tilde{x}, \tilde{y})  \|_{2}
       + G_{\ell} \Big( \frac{ (1+v) L_{h^*}^{ \frac{1}{v} } }{ 2v } \| \partial_{y} f(\tilde{x}, \tilde{y}) \|_{2} \Big)^{v}
       \\
=    & 
       \| \nabla_{x} f(\tilde{x}, \tilde{y})  \|_{2}
       + G_{\ell} \Big( \frac{ (1+v) }{ 2v } \Big)^{v} L_{h^*} \| \partial_{y} f(\tilde{x}, \tilde{y}) \|_{2}^{v}  ,
\end{align*}
where the last inequality is due to $(\varrho, p)$-uniformly convex of $f(\tilde{x}, y)$ in $y$ given $\tilde{x}$, i.e., (\ref{eq:uniformly_property}).
The first equality is due to Lemma \ref{lemma:Holder_to_uniformly_convex} that $\varrho = \frac{2v}{1+v} \cdot \frac{1}{ L_{h^*}^{ \frac{1}{v} } }$ and $p = 1 + \frac{1}{v}$.
\end{proof}

\section{Proof in Section \ref{sec:practical_section}}

\subsection{Proof of Theorem \ref{thm:outer_loop} }

\begin{proof}
Recall the notations $f^k_x(x) = g(x)  - y_k^{\top}\widetilde\ell(x) $, $f^k_y(y) = h(y) - y^{\top}\ell(x_{k+1}) $, where $\widetilde\ell(x) = \ell(x_k) + \nabla\ell(x_k)(x- x_k)$ for the case $\dom(h)\subseteq\mathbb R^m_+$ and  $\widetilde\ell(x) = \ell(x)$ for the case $\dom(h)\subseteq\mathbb R^m_-$. 

Define 
\begin{align*}
H_x^k(x) = f^k_x(x) + R^k_x(x), \quad H_y^k(y) = f^k_y(y) + R^k_y(y)
\end{align*}
and 
\begin{align*}
v_k = \arg\min_{x\in X}H_x^k(x) , \quad u_k = \arg\min_{y \in \dom(h)}H_y^k(y).
\end{align*}
Both of which are well-defined and unique due to the strong convexity of $H^k$.

Recall that a stochastic gradient of $f^k_x(x)$ can be computed by $\partial g(x; \xi_g) - \nabla \ell(x_k; \xi_\ell)^{\top}y_k$  for $\dom(h)\subseteq\mathbb R^m_+$ or $\partial g(x; \xi_g) - \nabla \ell(x; \xi_\ell)^{\top}y_k$ 
 for $\dom(h)\subseteq\mathbb R^m_-$. A stochastic gradient of $f^k_y(y)$ can be computed by $\partial h(y; \xi_h) - \ell(x_{k+1}; \xi_\ell')$, where $\xi_g, \xi_\ell, \xi_h, \xi_\ell'$ denote independent random variables.
 Then for $f^k_x(x)$ we have
 \begin{align*}
 \E[\| \nabla f^k_x(x)\|^2] = &\E[\|\partial g(x; \xi_g) - \nabla \ell(x_k; \xi_\ell)^{\top}y_k\|^2] \\
 \leq &2\E[\|\partial g(x; \xi_g) \|^2] +2 \E[\| \nabla \ell(x_k; \xi_\ell)^{\top}y_k\|^2]\\
  \leq &2\sigma^2 +2 \E[\| \nabla \ell(x_k; \xi_\ell)^{\top}y_k\|^2]\\
   \leq &2\sigma^2 +2 D^2\E[\| \nabla \ell(x_k; \xi_\ell)^{\top}\|^2]\\
   \leq &2\sigma^2 +2 D^2\sigma^2
 \end{align*}
 or
  \begin{align*}
 \E[\| \nabla f^k_x(x)\|^2] = &\E[\|\partial g(x; \xi_g) - \nabla \ell(x; \xi_\ell)^{\top}y_k\|^2] \\
 \leq &2\E[\|\partial g(x; \xi_g) \|^2] +2 \E[\| \nabla \ell(x; \xi_\ell)^{\top}y_k\|^2]\\
  \leq &2\sigma^2 +2 \E[\| \nabla \ell(x; \xi_\ell)^{\top}y_k\|^2]\\
    \leq &2\sigma^2 +2D^2 \E[\| \nabla \ell(x; \xi_\ell)^{\top}\|^2]\\
    \leq &2\sigma^2 +2 D^2\sigma^2
 \end{align*}
 where the second inequality uses Assumption~\ref{assumption:loss_function} (ii); the third inequality uses the assumption of $\max(\|y_k\|^2, \E[\|\ell(x_{k+1}; \xi)\|^2])\leq D^2$ for all $k\in\{1,\ldots, K\}$; the last inequality is due to Assumption~\ref{assumption:loss_function} (iii).
  For $f^k_y(y)$ we have
  \begin{align*}
 \E[\| \nabla f^k_y(y)\|^2] = &\E[\|\partial h(y; \xi_h) - \ell(x_{k+1}; \xi_\ell')\|^2] \\
 \leq &2\E[\|\partial h(y; \xi_h) \|^2] +2 \E[\| \ell(x_{k+1}; \xi_\ell')\|^2]\\
  \leq &2(\sigma^2 +D^2)
 \end{align*}
 where the second inequality uses Assumption~\ref{assumption:loss_function} (iv) and the assumption of $\max(\|y_k\|^2, \E[\|\ell(x_{k+1}; \xi)\|^2])\leq D^2$ for all $k\in\{1,\ldots, K\}$.
 We define a constant $G$, which will be used in our analysis:
 $$G := 17 \max\{2\sigma^2 +2 D^2\sigma^2, 2\sigma^2 +2 D^2 \}  ,$$
 which is in fact the role of $ 17 \sigma^2$ in the result of Proposition~\ref{thm:inner_loop_x}.

Next we could proceed to prove Theorem \ref{thm:outer_loop}.

Here we focus on the analysis using the convergence result in Proposition \ref{thm:inner_loop_x} corresponding to the non-smooth $f(z)$. Similar analysis can be done for using the result corresponding to smooth $f$. Applying  Proposition \ref{thm:inner_loop_x} to both $H^k_x$ and $H^k_y$ and adding their convergence bound together, we have
\begin{align}\label{eq:one_stage_x}
& \E[ H_{x}^{k}(x_{k+1}) + H_{y}^{k}(y_{k+1}) - H_{x}^{k}(v_{k}) - H_{y}^{k} (u_{k}) ]
\nonumber\\
\leq &
\frac{ G^2 }{\gamma ( T_{k}^{x} + 1 ) }
+ \frac{\gamma \E [\| x_{k} - v_{k} \|_{2}^{2} ] }{4 T_{k}^{x} ( T_{k}^{x} + 1 ) }
+ \frac{ G^2 }{\mu ( T_{k}^{y} + 1 ) }
+ \frac{\mu \E [ \| y_{k} - u_{k} \|_{2}^{2} ]  }{4 T_{k}^{y} ( T_{k}^{y} + 1 ) } .
\end{align}

The following inequalities hold due to the strong convexity of these two functions
$H_{x}^{k} (v_{k}) \leq H_{x}^{k} (x_{k}) - \frac{\gamma}{2} \E [ \| x_{k} - v_{k} \|^{2} ]$
and
$H_{y}^{k} (u_{k}) \leq H_{y}^{k} (y_{k}) - \frac{\mu}{2} \E [ \| y_{k} - u_{k} \|^{2} ]$.
Plug the above two inequalities to (\ref{eq:one_stage_x}),
\begin{align*}
& \E[ H_{x}^{k}(x_{k+1}) + H_{y}^{k}(y_{k+1}) - H_{x}^{k}(x_{k}) - H_{y}^{k} (y_{k}) ]
\nonumber\\
\leq &
\frac{ G^2}{\gamma ( T_{k}^{x} + 1 ) }
+ \frac{\gamma }{4 T_{k}^{x} ( T_{k}^{x} + 1 ) } \E [\| x_{k} - v_{k} \|_{2}^{2} ]  -  \frac{\gamma }{ 2 } \E [ \| x_{k} - v_{k} \|_{2}^{2} ]
\nonumber\\
&
+ \frac{ G^2 }{\mu ( T_{k}^{y} + 1 ) }
+ \frac{\mu }{4 T_{k}^{y} ( T_{k}^{y} + 1 ) } \E [ \| y_{k} - u_{k} \|_{2}^{2} ]  -  \frac{\mu }{ 2 } \E [ \| y_{k} - u_{k} \|_{2}^{2} ] .
\end{align*}
Recall the definition of $H_x^k(x) = f_x^k(x) + R_x^k(x) = f_x^k(x) + \frac{\gamma}{2} \| x - x_k \|^2$ and $H_y^k(y) = f_y^k (y) + R_y^k(y) = f_y^k (y) + \frac{\mu}{2} \| y - y_k \|$.
Since $T_{k}^{x} \geq 1$ and $T_{k}^{y} \geq 1$, we have
\begin{align*}
& \E[ f_{x}^{k}(x_{k+1}) +\frac{\gamma}{2}\|x_{k+1} - x_k\|^2+ f_{y}^{k}(y_{k+1}) + \frac{\mu}{2}\|y_{k+1} - y_k\|^2 - f_{x}^{k}(x_{k}) - f_{y}^{k} (y_{k}) ]
\nonumber\\
\leq &
\frac{ G^2}{\gamma ( T_{k}^{x} + 1 ) }
- \frac{\gamma }{ 4 } \E [ \| x_{k} - v_{k} \|_{2}^{2} ]
+ \frac{ G^2 }{\mu ( T_{k}^{y} + 1 ) }
- \frac{\mu }{ 4 } \E [ \| y_{k} - u_{k} \|_{2}^{2} ] .
\end{align*}
As a result, we have
\begin{align}\label{eq:one_stage_intermediate20}
&
\frac{1}{4} ( \E [ \gamma \| x_{k} -v_k \|^{2}  + \gamma \| x_{k+1} -x_k \|^{2} + \mu \| y_{k} - u_k \|^{2} +  \mu \| y_{k+1} - y_k \|^{2} ] )
\nonumber\\
\leq &
\E[ f_{x}^{k}(x_{k}) + f_{y}^{k} (y_{k}) - f_{x}^{k}(x_{k+1}) - f_{y}^{k} (y_{k+1}) +   \frac{ G^2}{\gamma ( T_{k}^{x} + 1 ) } + \frac{ G^2 }{\mu ( T_{k}^{y} + 1 ) }.
   \end{align}
   Let $T_k^x\geq k/\gamma + 1$ and  $T_k^y\geq k/\mu + 1$, we have
   \begin{align}\label{eq:one_stage_intermediate21}
   &
   \frac{1}{4} ( \E [ \gamma \| x_{k} -v_k \|^{2}  + \gamma \| x_{k+1} -x_k \|^{2} + \mu \| y_{k} - u_k \|^{2} +  \mu \| y_{k+1} - y_k \|^{2} ] )
   \nonumber\\
   \leq &
   \E[ \underbrace{   f_{x}^{k}(x_{k}) + f_{y}^{k} (y_{k}) - f_{x}^{k}(x_{k+1}) - f_{y}^{k} (y_{k+1})   }_{\text{\tiny{\circled{1}}}}]+   \frac{2 G^2}{k}
   \end{align}
   
   Let us consider the first term in the R.H.S of above inequality.   {For DC functions with $\dom(h)\subseteq\mathbb R^m_+$, recall $f^k_x(x) = g(x)  - y_k^{\top}(\ell(x_k) + \nabla\ell(x_k)(x - x_k)) $, $f^k_y(y) = h(y) - y^{\top}\ell(x_{k+1}) $. We have
       \begin{align*}
       &f^k_x(x_k) + f^k_y(y_k) - f^k_x(x_{k+1})  - f^k_y(y_{k+1})\\
       &= g(x_k) - y_k^{\top}\ell(x_k) + h(y_k) - y_k^{\top}\ell(x_{k+1}) - g(x_{k+1}) + y_k^{\top}(\ell(x_k) + \nabla\ell(x_k)(x_{k+1} - x_k)) \\
       & - h(y_{k+1})+ y_{k+1}^{\top}\ell(x_{k+1})\\
       &\leq g(x_k) + h(y_k)- y_k^{\top}\ell(x_k)  - (g(x_{k+1}) + h(y_{k+1}) -  y_{k+1}^{\top}\ell(x_{k+1})) \\
       &\leq f(x_k, y_k) - f(x_{k+1}, y_{k+1}),
       \end{align*}
       where we use $y_k\in\mathbb R^m_+$ and the convexity of $\ell(\cdot)$, i.e., $\ell(x_k) + \nabla\ell(x_k)(x_{k+1} - x_k)\leq \ell(x_{k+1})$.
       
       For Bi-convex functions, recall $f^k_x(x) = g(x)  - y_k^{\top}\ell(x)  $, $f^k_y(y) = h(y) - y^{\top}\ell(x_{k+1}) $. We have
       \begin{align*}
       &f^k_x(x_k) + f^k_y(y_k) - f^k_x(x_{k+1})  - f^k_y(y_{k+1})\\
       &= g(x_k) - y_k^{\top}\ell(x_k) + h(y_k) - y_k^{\top}\ell(x_{k+1}) - g(x_{k+1}) + y_k^{\top}\ell(x_{k+1})  - h(y_{k+1}) + y_{k+1}^{\top}\ell(x_{k+1})\\
       &=g(x_k) + h(y_k)- y_k^{\top}\ell(x_k)  - (g(x_{k+1}) + h(y_{k+1}) -  y_{k+1}^{\top}\ell(x_{k+1})) \\
       &= f(x_k, y_k) - f(x_{k+1}, y_{k+1}).
       \end{align*}
   }
   Hence, we have
   \begin{align}\label{eq:one_stage_intermediate2}
   &
   \frac{1}{4} ( \E [ \gamma \| x_{k} -v_k \|^{2}  + \gamma \| x_{k+1} -x_k \|^{2} + \mu \| y_{k} - u_k \|^{2} +  \mu \| y_{k+1} - y_k \|^{2} ] )
   \nonumber\\
   \leq &
   \E[ f(x_{k}, y_{k}) - f(x_{k+1}, y_{k+1})  ]+   \frac{2 G^2}{k}
   \end{align}
   
   Next, we can bound the sequence of $x_k$ and $y_k$ separately. Let us focus on the sequence of $x_k$   and the analysis for the sequence of $y_k$ is  similar.
   \begin{align}\label{eq:one_stage_intermediate2}
   \frac{1}{4} \E [ \gamma \| x_{k} -v_k \|^{2}  + \gamma \| x_{k+1} -x_k \|^{2} ]
   \leq &
   \E[ f(x_{k}, y_{k}) - f(x_{k+1}, y_{k+1})  ]+   \frac{2 G^2}{k}
   \end{align}
   
   Next dividing $\gamma$ and then multiplying $\omega_{k}$ and  on both sides and taking summation over $k = 1, ..., K$ where $\alpha \geq 1$, one has
   \begin{align}\label{eq:outer_loop_upperbound}
   &
   \frac{ 1}{ 4 }
   \E\bigg[ \sum_{k=1}^{K} \omega_{k} (  \| x_{k} -v_k \|^{2}  + \| x_{k+1} -x_k \|^{2})\bigg]
   \nonumber\\
   \leq &
   \sum_{k=1}^{K} \frac{\omega_{k}}{\gamma} \E[ f(x_{k}, y_{k}) - f(x_{k+1}, y_{k+1}) ]
   + 2G^2 \sum_{k=1}^{K} \frac{ \omega_{k} }{ k\gamma }     .
   \end{align}
   
   For the LHS of (\ref{eq:outer_loop_upperbound}), we have 
   \begin{align*}
     \E [ \| x_{\tau} - v_\tau \|^{2}  + \| x_{\tau+1} - x_\tau \|^{2} ]
     = \frac{ \sum_{k=1}^{K} \omega_k \E [ \| x_{k} - v_k \|^{2}  + \| x_{k+1} - x_k \|^{2} ] }{ \sum_{k=1}^{K} \omega_k }   ,
   \end{align*}
   where $\tau$ is sampled by $P(\tau = k) = \frac{ k^\alpha }{ \sum_{s=1}^K s^\alpha }$.
   
   For the RHS of (\ref{eq:outer_loop_upperbound}), let us consider the first term.
   According to the setting $\omega_k = k^\alpha$ with $\alpha\geq 1$ and following the similar analysis of Theorem 2 in~\citep{chen18stagewise}, we have
   \begin{align*}
       & \sum_{k=1}^L \omega_k ( f(x_k, y_k) - f(x_{k+1}, y_{k+1}) )
       \\
     = & \sum_{k=1}^K ( \omega_{k-1} f(x_k, y_k) - \omega_k f(x_{k+1}, y_{k+1}) ) 
       + \sum_{k=1}^K ( \omega_k - \omega_{k-1} ) f(x_{k, y_{k}})
       \\
     = & \omega_0 f(x_1, y_1) - \omega_K f(x_{K+1}, y_{K+1}) + \sum_{k=1}^K ( \omega_k - \omega_{k-1} ) f(x_{k, y_{k}})
       \\
     = & \sum_{k=1}^K ( \omega_k - \omega_{k-1} ) ( f(x_k , y_k) - f(x_{k+1}, y_{k+1}) ) 
       \\
     \leq & \sum_{k=1}^K ( \omega_k - \omega_{k-1} ) M  
     =      M \omega_K 
     =      M K^\alpha    ,
   \end{align*}
   where the third equality is due to $\omega_0 = 0$ and the inequality is due to Assumption \ref{assumption:loss_function} (v).
   Then for the second term of RHS of (\ref{eq:outer_loop_upperbound}),
   \begin{align*}
     \sum_{k=1}^K \frac{\omega_k}{k} = \sum_{k=1}^K k^{\alpha - 1}
     \leq K K^{\alpha - 1}
     =    K^{\alpha}    .
   \end{align*}
   
   Plugging the above three terms back into (\ref{eq:outer_loop_upperbound}) and dividing both sides by $\sum_{k=1}^K \omega_k$,
   we have
   \begin{align}\label{eq:bound_x}
   \E[  \| x_{\tau} -v_\tau \|^{2}  + \| x_{\tau+1} -x_\tau \|^{2}]\leq   &
   \frac{ 4(M+2G^2) (\alpha + 1) }{ \gamma K }   ,
   \end{align}
   due to $\sum_{k=1}^K k^\alpha \geq \int_0^K s^\alpha d s = \frac{K}{\alpha + 1}$.
   
   Similarly, by setting $\omega_k = k^\alpha$ with $\alpha\geq 1$ and following the similar analysis of Theorem 2 in~\citep{chen18stagewise},  we have
   \begin{align}\label{eq:bound_y}
   \E[  \| y_{\tau} -u_\tau \|^{2}  + \| y_{\tau+1} -y_\tau \|^{2}]\leq    &
   \frac{ 4(M+2G^2) (\alpha + 1) }{\mu K }   .
   \end{align}
   In addition, we have
   \begin{align*}
   \E[  \| x_{\tau+1} -v_\tau \|^{2}  ]\leq 2\E[  \| x_{\tau} -v_\tau \|^{2}  + \| x_{\tau+1} -x_\tau \|^{2}]\leq    &\frac{ 8(M+2G^2) (\alpha + 1) }{\gamma K }, \\
   \E[  \| y_{\tau+1} -u_\tau \|^{2}  ]\leq 2\E[  \| y_{\tau} -u_\tau \|^{2}  + \| y_{\tau+1} -u_\tau \|^{2}]\leq    &\frac{ 8(M+2G^2) (\alpha + 1) }{\mu K } .
   \end{align*}

\end{proof}

\subsection{Proof of Proposition~\ref{thm:inner_loop_x}}
\begin{proof}
This proof is similar to the proof of Proposition 2 in~\cite{xu2018stochastic}. For completeness, we include it here.

{\bf Smooth Case.} When $f(z)$ is $L$-smooth and $R(z)$ is $\gamma$-stronglly convex, we then first have the following lemma from~\citep{DBLP:conf/icml/ZhaoZ15}. 
\begin{lem}\label{lem:for:prop2}
Under the same assumptions in Proposition~\ref{thm:inner_loop_x}, we have
\begin{align*}
\E[H(z_{t+1}) - H(z)] \leq & \frac{\|z_t - z\|^2}{2\eta_t} - \frac{\|z- z_{t+1} \|^2}{2\eta_t} - \frac{\gamma}{2}\|z - z_{t+1}\|^2  + \eta_t\sigma^2.
\end{align*}
\end{lem}
The proof of this lemma is similar to the analysis to proof of Lemma 1 in~\citep{DBLP:conf/icml/ZhaoZ15}. Its proof can be found in the analysis of Lemma 7 in~\cite{xu2018stochastic}.

Let us set $w_t = t$, then by Lemma~\ref{lem:for:prop2} we have
\begin{align*}
&\sum_{t=1}^Tw_{t+1}(H(z_{t+1})  - H(z))\\
\leq& \sum_{t=1}^T\left(\frac{w_{t+1}}{2\eta_t}\|z - z_t\|^2 - \frac{w_{t+1}}{2\eta_t}\|z - z_{t+1}\|^2 -\frac{\gamma w_{t+1}}{2}\|z - z_{t+1}\|^2  \right) + \sum_{t=1}^T\eta_t w_{t+1} \sigma^2\\
\leq & \sum_{t=1}^T\bigg(  \frac{w_{t+1}}{2\eta_t}  - \frac{w_{t}}{2\eta_{t-1}}- \frac{\gamma w_{t}}{2}\bigg)\|z - z_t\|^2 + \frac{w_1/\eta_0 + \gamma w_1}{2} \|z - z_1\|^2   + \sum_{t=1}^T\eta_t w_{t+1} \sigma^2\\
\leq & \frac{2\gamma}{3} \|z - z_1\|^2  + \sum_{t=1}^T\frac{3\sigma^2}{\gamma},
\end{align*}
where the last inequality is due to the settings of $\eta_t$ and $w_t$ such that $ \frac{w_{t+1}}{\eta_t}  - \frac{w_{t}}{\eta_{t-1}}- \gamma w_{t} =\frac{\gamma(t+1)^2}{3} - \frac{\gamma t^2}{3} - \gamma t = \frac{\gamma(1-t)}{3} \leq 0$. Then by the convexity of $H = f+R$ and the update of $\hat z_T$, we know
\begin{align*}
H(\hat z_T)  - H(z)\leq \frac{4\gamma\|z - z_1\|^2}{3T(T+3)} + \frac{6\sigma^2}{(T+3)\gamma}.
\end{align*}
We complete the proof of smooth case by letting $z=z_*$ in above inequality.

{\bf Non-smooth Case.} We then consider the case of $f(z)$ is non-smooth. Recall that the update of $z_{t+1}$ is 
\begin{align*}
z_{t+1} = \arg\min_{z \in \Omega} \partial f(z_t; \xi_t)^{\top}z  + R(z) + \frac{1 }{2\eta_t} \|z - z_t\|^2.
\end{align*}
By the optimality condition of $z_{t+1}$ and the strong convexity of above objective function, we know for any $z\in\Omega$,
\begin{align*}
&\partial f(z_t; \xi_t)^{\top}z+ R(z) +   \frac{1}{2\eta_t}\|z- z_{t}\|^2\\
\geq &\partial f(z_t; \xi_t)^{\top}z_{t+1} +R(z_{t+1}) + \frac{1}{2\eta_t}\|z_{t+1}- z_{t}\|^2 + \frac{1/\eta_t + \gamma}{2}\|z - z_{t+1}\|^2,
\end{align*}
which implies
\begin{align*}
&\partial f(z_t; \xi_t)^\top (z_t - z)+ R(z_{t+1})   - R(z)\\
\leq & (z_t - z_{t+1})^{\top}\partial f(z_t; \xi_t)- \frac{1}{2\eta_t}\|z_{t+1}- z_{t}\|^2 +     \frac{1}{2\eta_t}\|z- z_{t}\|^2 -  \frac{1/\eta_t + \gamma}{2}\|z - z_{t+1}\|^2\\
\leq &\frac{\eta_t \|\partial f(z_t; \xi_t)\|^2}{2}+     \frac{1}{2\eta_t}\|z- z_{t}\|^2 -  \frac{1/\eta_t + \gamma}{2}\|z - z_{t+1}\|^2.
\end{align*}
Taking expectation on both sides of above inequality and using the convexity of $f(z)$, then we get
\begin{align*}
&\E[f(z_{t}) - f(z)+ R(z_{t+1}) - R(z)]\\
\leq &\frac{\eta_t\E[ \|\partial f(z_t; \xi_t)\|^2]}{2}+\E\bigg[\frac{1}{2\eta_t}\|z- z_{t}\|^2 -  \frac{1/\eta_t + \gamma}{2}\|z - z_{t+1}\|^2\bigg]\\
\leq &\frac{\eta_t\sigma^2}{2}+\E\bigg[\frac{1}{2\eta_t}\|z- z_{t}\|^2 -  \frac{1/\eta_t + \gamma}{2}\|z - z_{t+1}\|^2\bigg].
\end{align*}
Multiplying both sides of above inequality by $w_{t} = t$ and taking summation over $t=1,\ldots, T$, then
\begin{align*}
&\E\bigg[\sum_{t=1}^Tw_{t}(f(z_t) - f(z)+ R(z_{t+1}) - R(z))\bigg]\\
&\leq \sum_{t=1}^T2\sigma^2w_{t}\eta_t+ \E\bigg[\sum_{t=1}^T \frac{w_{t}}{2\eta_t}\|z- z_{t}\|^2 -  \frac{w_{t}/\eta_t + w_{t}\gamma}{2}\|z - z_{t+1}\|^2\bigg].
\end{align*}
We rewrite above inequality, then
\begin{align*}
&\E\bigg[\sum_{t=1}^Tw_{t}(f(z_{t}) - f(z)+ R (z_{t})   - R(z))\bigg]\\
\leq&\E\bigg[\sum_{t=1}^Tw_{t}(R(z_{t}) - R(z_{t+1}))\bigg]+ \sum_{t=1}^T2\sigma^2w_{t}\eta_t+    \E\bigg[\sum_{t=1}^T \left(\frac{w_{t}}{2\eta_t}-\frac{w_{t-1}/\eta_{t-1} + w_{t-1}\gamma}{2}\right)\|z- z_{t}\|^2\bigg]\\
\leq&\E\bigg[\sum_{t=1}^Tw_{t}(R(z_{t}) - R(z_{t+1}))\bigg]+ \frac{8\sigma^2 T}{\gamma}+ \frac{\gamma\|z - z_1\|^2}{8},
\end{align*}
where the last inequality is due to $w_t =t, \eta_t = 4/(\gamma t)$, and $\frac{w_{t}}{2\eta_t}-\frac{w_{t-1}/\eta_{t-1} + w_{t-1}\gamma}{2}\leq 0, \forall t\geq 2$.
The let us consider the first term, we have
\begin{align}\label{prop2:ineq1}
\nonumber &\E\bigg[\sum_{t=1}^Tw_{t}(f(z_{t}) - f(z)+ R (z_{t})   - R(z))\bigg]\\
\nonumber \leq &w_0R(z_1) - w_{T}R(z_{T+1}) + \E\bigg[\sum_{t=1}^T(w_{t} - w_{t-1})R(z_{t})\bigg]+ \frac{8\sigma^2 T}{\gamma}+ \frac{\gamma\|z - z_1\|^2}{8}\\
=  &\E\bigg[\sum_{t=1}^T(w_{t} - w_{t-1})R(z_{t})\bigg]+ \frac{8\sigma^2 T}{\gamma}+ \frac{\gamma\|z - z_1\|^2}{8}.
\end{align}
Next, we want to show for any $z_t$ we have 
\begin{align}\label{prop2:bounded:dist}
\E[\|z_t - z_1\|^2]\leq \frac{\sigma^2}{\gamma^2}.
\end{align} 
We prove it by induction. It is easy to show that the inequality (\ref{prop2:bounded:dist}) holds for $t=1$. We then aussume the inequality (\ref{prop2:bounded:dist}) holds for $t$. By the update of $z_{t+1}=\arg\min_{z\in\Omega} \partial f(z_t; \xi_t)^{\top}z  + \frac{\gamma}{2}\|z-z_1\|^2 + \frac{1 }{2\eta_t} \|z - z_t\|^2= \arg\min_{z\in\Omega} \frac{1}{2}\|z-z_{t+1}'\|^2$, where $z_{t+1}' = \frac{\gamma z_1 + \frac{1}{\eta_t}z_t - \partial f(z_t; \xi_t)}{\gamma + \frac{1}{\eta_t}}$. Then 
\begin{align*}
\E[\|z_{t+1} - z_1\|^2]\leq & \E[\|z_{t+1}'-z_1\|^2] 
=  \frac{1}{(\gamma + \frac{1}{\eta_t})^2}\E\bigg[\bigg\| \frac{z_t -z_1}{\eta_t}- \partial f(z_t; \xi_t) \bigg\|^2\bigg]\\
\leq &\frac{1}{(\gamma + \frac{1}{\eta_t})^2}\bigg( \frac{1+\gamma\eta_t}{\eta_t^2}\E[\| z_t -z_1\|^2] + \bigg(1+\frac{1}{\gamma\eta_t}\bigg)\E[\|\partial f(z_t; \xi_t)\|^2] \bigg)\\
\leq &\frac{1}{(\gamma + \frac{1}{\eta_t})^2}\bigg( \frac{(1+\gamma\eta_t)\sigma^2}{\eta_t^2\gamma^2} + \bigg(1+\frac{1}{\gamma\eta_t}\bigg)\sigma^2\bigg)  = \frac{\sigma^2}{\gamma^2}.
\end{align*} 
Then by induction we know the inequality (\ref{prop2:bounded:dist}) holds for all $t\geq 1$.
Combining inequalities (\ref{prop2:ineq1}) and (\ref{prop2:bounded:dist}) we get
\begin{align*}
&\E\bigg[\sum_{t=1}^Tw_{t}(f(z_{t}) - f(z)+ R (z_{t})   - R(z))\bigg]\\
\leq & \E\bigg[\sum_{t=1}^T(w_{t} - w_{t-1})\frac{\sigma^2}{2\gamma}\bigg]+ \frac{8\sigma^2 T}{\gamma}+ \frac{\gamma\|z - z_1\|^2}{8}\\
= &\frac{17\sigma^2 T}{2\gamma}+ \frac{\gamma\|z - z_1\|^2}{8}.
\end{align*}
Then by the convexity of $H = f+R$ and the update of $\hat z_T$, we know
\begin{align*}
\E\bigg[H(\hat z_T) - H(z)\bigg]\leq \frac{17\sigma^2}{\gamma (T+1)}+ \frac{\gamma\|z - z_1\|^2}{4T(T+1)}.
\end{align*}
We complete the proof of non-smooth case by letting $z=z_*$ in above inequality.
\end{proof}

\subsection{Proof of Lemma \ref{lem:epsilon_critical_smooth_g}}
 
\paragraph{Part I.}
We consider $g$ is $L_g$-smooth and $\ell$ is $G_\ell$-Lipschitz continuous.
Due to the first order optimality of $f_{x}^{k} (x)$ at $v_{k}$ (and smoothness of $f_{x}^{k} (x)$),
\begin{align*}
0 =   &
\nabla g(v_{k}) - \nabla \ell(x_{k}) y_{k} + \gamma (v_{k} - x_{k})
\\
=     &
\nabla g(x_{k}) - \nabla \ell(x_{k}) y^*(x_{k})
+ \gamma (v_{k} - x_{k})
\\
&
+ \nabla g(v_{k}) - \nabla g(x_{k})
+ \nabla \ell(x_{k}) ( y^*(x_{k}) - y_{k} )
\\
=     &
\nabla F(x_{k})
+ \gamma (v_{k} - x_{k})
\\
&
+ \nabla g(v_{k}) - \nabla g(x_{k})
+ \nabla \ell(x_{k})  ( y^*(x_{k}) - y_{k} )    ,
\end{align*}
where $y^*(x_k) = \arg\min_{y \in dom(h)} h(y) - \ell(x_k)^{\top} y$.
Let $\tilde f_{y}^{k} (y) = h(y) - \ell(x_{k})^{\top} y$.
The second equality is due to Theorem 10.13 of \cite{rockafellar2009variational} and the uniqueness of $y^*(x_{k})$ ($h$ is uniformly convex).

To bound $\| \nabla F(x_{k}) \|$ we have,
\begin{align*}
\| \nabla F(x_{k}) \|
\leq &
\gamma \| v_{k} - x_{k} \|
+ \| \nabla g(v_{k}) - \nabla g(x_{k}) \|
+ \| \nabla \ell(x_{k}) \| \cdot \| y^*(x_{k}) - u_{k} + u_{k} - y_{k} \|
\\
\leq& \gamma \| v_{k} - x_{k} \|
+ L_{g} \| v_{k} - x_{k} \|
+ G_{\ell} \underbrace{ \| y^*(x_{k}) - u_{k} \| }_{ \text{\tiny{\circled{1}}} }
+ G_{\ell} \| u_{k} - y_{k} \|   .
\end{align*}

To handle {\tiny{\circled{1}}},
we could use the $(\varrho, p)$-uniform convexity of $\tilde f_{y}^{k}$ (since $\nabla h^*$ is assumed to be $(L_{h^*}, v)$-H\"older continuous) as follows
\begin{align*}
       \| y^*(x_{k}) - u_{k} \|^{p-1}
\leq &
       \frac{1}{\varrho} \| \partial \tilde f^k_y(y^*(x_{k})) - \partial \tilde f^k_y(u_{k}) \|
       \\
=    &
       \frac{1}{\varrho} \|  \partial  h(u_{k})  - \ell(x_{k}) ) \|
       \\
\leq &
       \frac{1}{\varrho} ( \|  \partial  h(u_{k})  - \ell(x_{k+1})  \| +  \|  \ell(x_{k+1})  - \ell(x_{k})  \| )
\\
\leq &
       \frac{1}{\varrho} \| \mu ( u_{k} - y_{k} ) \|   + \frac{G_\ell}{\varrho}\|x_{k+1} - x_k\| ,
\end{align*}
where the first inequality is due to (\ref{eq:uniformly_property}).
The first equality is due to the first order optimality of $\tilde f_{y}^{k} (y)$ at $y^*(x_{k})$, i.e., $ 0 \in \partial \tilde f_{y}^{k} (y^*(x_{k}))$.
The third  inequality is due to the first order optimality of $f_{y}^{k}(y) + R^k_y(y)$ at $u_{k}$, i.e.,
$0 \in \partial  h(u_{k})  - \ell(x_{k+1}) + \mu ( u_{k} - y_{k} )$.

Since $\varrho = \frac{ 2v }{ 1+v } \Big( \frac{1}{L_{h^*}} \Big)^{\frac{1}{v}}$ and $v = \frac{1}{p-1}$ (Lemma \ref{lemma:Holder_to_uniformly_convex}), one has
$\| y^*(x_{k}) - u_{k} \| \leq \mu^{v} \Big( \frac{1+v}{2v} \Big)^{v} L_{h^*} \| u_{k} - y_{k} \|^{v}+  G_\ell^{v} \Big( \frac{1+v}{2v} \Big)^{v} L_{h^*} \| u_{k} - y_{k} \|^{v}$.
Therefore,
\begin{align*}
\| \nabla F(x_{k}) \|
\leq
&
\gamma \| v_{k} - x_{k} \|
+ L_{g} \| v_{k} - x_{k} \|
\\
&
+ G_{\ell} \mu^{v} \Big( \frac{1+v}{2v} \Big)^{v} L_{h^*} \| u_{k} - y_{k} \|^{v} +  G_{\ell}^{v+1} \Big( \frac{1+v}{2v} \Big)^{v} L_{h^*} \| x_{k+1} - x_{k} \|^{v}
+ G_{\ell} \| u_{k} - y_{k} \|     .
\end{align*}

\paragraph{Part II.}
We consider $g$ is non-smooth and $\ell$ is $G_\ell$-Lipschitz continuous and $L_\ell$-smooth and $\max_{y \in dom(h)} \| y \| \leq D$.
Due to the first order optimality of $f_{x}^{k}$ at $v_{k}$,
\begin{align*}
0 \in &
\partial g(v_{k}) - \nabla \ell(x_{k})y_{k}  + \gamma ( v_{k} - x_{k} )
\\
=     &
\partial g(v_{k}) -\nabla \ell(v_{k}) y^*(v_{k})
+ \gamma ( v_{k} - x_{k} )
+ \nabla \ell(v_{k})y^*(v_{k})  -\nabla \ell(x_{k}) y_{k}
\\
=     &
\partial F(v_{k})
+ \gamma ( v_{k} - x_{k} )
+  \nabla \ell(v_{k})y^*(v_{k}) - \nabla \ell(v_{k}) y_{k}
+  \nabla \ell(v_{k})y_{k} -  \nabla \ell(x_{k})y_{k}
\\
=     &
\partial F(v_{k})
+ \gamma ( v_{k} - x_{k} )
+ \nabla \ell(v_{k}) ( y^*(v_{k}) - y_{k} )
+( \nabla \ell(v_{k}) - \nabla \ell(x_{k}) )  y_{k}   .
\end{align*}
The second equality is due to Theorem 10.13 of \cite{rockafellar2009variational} and the uniqueness of $y^*(v_{k})$ ($h$ is uniformly convex).

Therefore, by $G_{\ell}$-Lipschits continuity of $\ell$, $L_{\ell}$-smoothness of $\ell$ and $\max_{y \in \dom(h)} \| y \| \leq D_{y}$,
\begin{align*}
& \dist(0, \partial F(v_{k}))
\\
\leq &
\gamma \| v_{k} - x_{k} \|
+ G_{\ell} \| y^*(v_{k}) - y_{k} \|
+ D_{y} L_{\ell} \| v_{k} - x_{k} \|
\\
\leq &
\gamma \| v_{k} - x_{k} \|
+ G_{\ell} \| u_{k} - y_{k} \|
+ G_{\ell} \underbrace{ \| y^*(v_{k}) - u_{k} \| }_{\text{\tiny{\circled{1}}}}
+ D_{y} L_{\ell} \| v_{k} - x_{k} \|   .
\end{align*}

To deal with {\tiny{\circled{1}}}, one could employ $(\varrho, p)$-uniform convexity of $\tilde{f}_{y}^{k} = h(y)  - \ell(v_{k})^{\top} y$,
\begin{align*}
\| y^*(v_{k}) - u_{k} \|^{p-1}
\leq &
\frac{1}{\varrho} \|\partial h(u_{k})  - \ell(v_{k}) \|
\\
\leq &
\frac{1}{\varrho} \| \partial  h(u_{k})  - \ell(x_{k+1}) \| + \| \ell(x_{k+1}) - \ell(v_{k}) \|
\\
\leq &
\frac{\mu}{\varrho} \| u_{k} - y_{k} \| + \frac{G_{\ell}}{\varrho} \| x_{k+1} - v_{k} \|  ,
\end{align*}
where the first inequality is due to $(\varrho, p)$-uniform convexity of $h$ and the first order optimality of $\tilde{f}_y^k$ at $y^*(v_{k})$.
The last inequality is due to the first order optimality of $f_{y}^{k} + R_y^k$ at $u_{k}$,
i.e., $0 \in \partial  h(u_{k}) - \ell(x_{k+1}) + \mu ( u_{k} - y_{k} )$,
and $G_{\ell}$-Lipschits continuity of $\ell$.

Since $\nabla h^*$ is $(L_{h^*}, v)$-H\"older continuous by assumption, by Lemma \ref{lemma:Holder_to_uniformly_convex}, $\varrho = \frac{2v}{1+v} \Big( \frac{1}{L_{h^*}} \Big)^{ \frac{1}{v} }$.
Then one has
\begin{align*}
\| y^*(v_{k}) - u_{k} \|
\leq &
\Big( \frac{1+v}{2v} \Big)^{v} L_{h^*} \Big( \mu \| u_{k} - y_{k} \| + G_{\ell} \| x_{k+1} - v_{k} \|  \Big)^{v}   .
\end{align*}

Therefore, one has
\begin{align*}
& \dist(0, \partial F(v_{k}))
\\
\leq &
\gamma \| v_{k} - x_{k} \|
+ G_{\ell} \| u_{k} - y_{k} \|
\\
&
+ G_{\ell} \Big( \frac{1+v}{2v} \Big)^{v} L_{h^*} \Big( \mu \| u_{k} - y_{k} \| + G_{\ell} \| x_{k+1} - v_{k} \|  \Big)^{v}
+ D L_{\ell} \| v_{k} - x_{k} \|    .
\end{align*}

\end{document}